# Hopes and Fears – Emotion Distribution in the Topic Landscape of Finnish Parliamentary Speech 2000-2020

Anna Ristilä[1], Otto Tarkka[1], Veronika Laippala[1], Kimmo Elo[2]

[1]University of Turku, [2]University of Eastern Finland

## Abstract

Existing research often treats parliamentary discourse as a homogeneous whole, overlooking topic-specific patterns. Parliamentary speeches address a wide range of topics, some of which evoke stronger emotions than others. While everyone has intuitive assumptions about what the most emotive topics in a parliament may be, there has been little research into the emotions typically linked to different topics. This paper strives to fill this gap by examining emotion expression among the topics of parliamentary speeches delivered in *Eduskunta*, the Finnish Parliament, between 2000 and 2020. An emotion analysis model is used to investigate emotion expression in topics, from both synchronic and diachronic perspectives. The results strengthen evidence of increasing positivity in parliamentary speech and provide further insights into topic-specific emotion expression within parliamentary debate.

## Introduction

Parliamentary debate is a central arena for democratic communication where language is not only used to neutrally inform but is also infused with emotion to persuade, justify, and legitimize. In recent years, the emotional dimensions of political discourse have attracted growing scholarly attention in the fields of political communication and digital humanities. Despite the interest, relatively little is known about how emotions are distributed across different topics within parliamentary debate, or if certain types of events, such as national crises or major changes in the parliamentary power balance, affect the debate's emotion levels. Studying how a parliament expresses emotion around certain topics or issues is important because it both reflects underlying societal values regarding those issues and influences the wider public debate about them.

Most of the existing parliamentary research has focused on English-language parliaments, partly due to issues of data availability. These parliaments are typically characterized by two-party systems. However, new datasets have become available and enabled increased variation in the targets studied. Parliaments composed of multiple parties represent a broader spectrum of political views and reflect the values present in society with greater resolution than a one or two-party system. The combination of multiparty variation and institutional consistency makes the corpus of *Eduskunta*, the Finnish Parliament, particularly well suited for large-scale exploration of emotional variation in a political context.

This study addresses the research gap related to topic-emotion interactions in a non-Anglophone parliamentary context by examining how emotions are expressed across different topics in Finnish parliamentary speeches between 2000 and 2020. By combining emotion detection with topic modeling, it paints a topic landscape of how emotion varies by policy areas and over time. Close reading is used to complement computational analyses. The expression rates of various emotions are interpreted as indicators of average societal values and stances towards certain themes common to parliamentary discussions.

While earlier work on affect in politics has typically concentrated on simple positive-negative dimensions (see e.g. Rheault *et al.* 2016, Abercrombie and Batista-Navarro 2020), single events in time, or voting patterns, few studies have attempted to conceptualize the broader structure of how topics relate to emotions within parliamentary speech. This paper's approach is *topic landscape* (earlier used in Ristilä 2025), a framework for examining the distribution of topics in relation to variables of interest, in this case emotions. Rather than focusing on topic extraction as such, the topic landscape approach emphasizes mapping the shape of thematic variation across the corpus and identifying where emotions concentrate or disperse within that structure.

This paper addresses the following research questions:

- How does the topic landscape of *Eduskunta* look like when examined through the lens of emotion?
- How has emotion expression evolved over time in *Eduskunta*?
- What kind of wider societal developments do the results imply?

The paper contributes to research on political communication by providing new evidence on the affective dimensions of topics typical to parliamentary contexts and extends emotion research beyond major English-language parliaments to a Nordic, bilingual context. It also offers insights into how emotion expression has evolved and how some parliamentary events may affect emotion expression in politics. Furthermore, it contributes to corpus-based studies by proposing a definition for *topic-landscape approach* which uses topics as an organizing frame for analyzing large parliamentary corpora while explicitly linking this structure to other variables such as emotion expression.

## Background

Finland has a unicameral parliamentary system, with legislative power vested in *Eduskunta*, a 200-member parliament elected every four years through open-list proportional representation. The system is characterized by multi-party coalition governments, which are the norm rather than the exception, and by a strong committee structure that prepares legislation before it reaches plenary debate. Plenary sessions follow a predetermined agenda (*asialista*) set by the Speaker's Council, meaning that individual Members of Parliament (MPs) have limited influence over what topics are discussed, and exercise their agency primarily in how they address the issues on the agenda. This institutional structure shapes the linguistic environment of parliamentary debate.

The following subsections cover several terms commonly related to emotion research, discuss earlier research, and introduce some key concepts that are used for interpreting the results.

### Terminology

Earlier research on emotions has been extensive and filled with conflicting terminology. As Du Bois and Kärkkäinen (2012:434) have summarized, "[t]he display of emotion (…) can vary from blatant to subtle to unspoken, or anything in between", which has made the study of emotion an interesting but difficult subject to talk about. Scholars have referred to related phenomena as *emotion*, *sentiment*, *valence*, *affect*, *evaluation*, *attitude*, *appraisal*, or *stance*, often defining them in partially overlapping ways. This conceptual plurality reflects the interdisciplinary nature of emotion research, drawing from psychology, linguistics, anthropology, and sociology.

In linguistics, the term *affect* is often used to capture the interactional dimensions of emotion as expressed through language (e.g. Du Bois & Kärkkäinen 2012), while *evaluation* tends to emphasize judgments of value or worth. For example, Sanger (2020) uses the term *affect* to talk about legislation related to abortions. She describes how *affect* has an influence on the *evaluation* of the subject, which influences the legislation, which again influences public evaluation on the subject, which in turn influences the way women are expected to feel about abortion. The concepts of affective and evaluative language have been situated within a broader model of stance by Du Bois (2007), so that the affective orientation forms just one component of how speakers position themselves toward the content of their talk. Stance has been applied, for example, by Cheng *et al.* (2024) in the study of data protection laws. This development represented a shift away from treating emotion merely as an internal psychological state and toward understanding it as a socially distributed, linguistically mediated practice.

While both affect and evaluation are central to political stance-taking, the current analysis concentrates on affect, rather than on explicit evaluative or judgmental language, and sees affect as part of stance. Here affect is defined as *active linguistic expression of emotion triggered by contextual factors*, while stance is defined as *a combination of affects, evaluations, and positions toward the topics under discussion*.

Because sentiment analysis has been a widespread method in digital language research, it is also necessary to define the distinction between sentiments and emotions. *Sentiments* are considered here as the basic positive, neutral, and negative categories, while *emotions* are seen as the more nuanced categories with specific names such as 'hope', 'fear', or 'disgust'. An example of a multilingual sentiment analysis model is Barbieri *et al.* (2022) and another of an English language emotion analysis model is Hartmann (2022). While this paper will mostly talk about emotion expression, when talking about e.g. all positive emotions together as a group we may refer to the grouping as a 'positive sentiment'. However, coarse-grained positive–negative classification approaches can obscure crucial distinctions in how speakers position the topics they speak of, which is why this study prioritizes emotion-level analysis while using sentiment groupings and a combined average sentiment score mostly for comparability with earlier work.

## Stance and emotions

In this study, *emotions* are interpreted as partial indications of *stance*. This subsection further elaborates these two concepts and their relationship.

The concept of stance has been approached by several disciplinary traditions. Early anthropological work, such as Ochs and Schieffelin (1989), saw stance as the intersection of affective, epistemic, and social positioning, which already places emotions (i.e. affect) as a key part of stance. In corpus linguistics, Conrad and Biber (2000; see also Biber *et al.* 1999) identified grammatical devices associated with stance and distinguished three broad categories: epistemic, attitudinal, and style stance. Epistemic stance concerned the degree of certainty, doubt, or commitment to truth, typically realized through modals, evidentials, and related markers. Attitudinal stance expressed feelings, attitudes, and value judgments, often through evaluative adjectives, adverbs, or verbs of affect. Style stance captured the speaker's rhetorical manner of delivery, such as hedging, intensification, or metalinguistic commentary. Among Biber and Conrad's three categories, attitudinal stance is most directly concerned with the emotions central to this study. It covers affects toward debated topics and participants, and highlights stance as a socially situated act, an important perspective in parliamentary contexts.

Around the same time as Biber, Hyland (1999; 2005) developed a model of stance in academic writing, focusing on metadiscursive features such as hedges, boosters, self-mentions, and attitude markers, and showing how authors construct credibility and alignment with readers. Du Bois' (2007) stance triangle highlighted an interactional perspective, conceptualizing stance as the simultaneous acts of evaluation, positioning, and alignment. In more detail, Du Bois (2007:163) defines stance as "a public act by a social actor, achieved dialogically through overt communicative means, of simultaneously evaluating objects, positioning subjects (self and others), and aligning with other subjects, with respect to any salient dimension of the sociocultural field." Subsequent work (Du Bois & Kärkkäinen 2012) has extended Du Bois' framework to include the affective dimensions of stance-taking, emphasizing how emotional orientation contributes to evaluation and alignment in interaction.

Based on the theoretical literature described above, stance is seen here to emerge mainly from both affective and evaluative responses to topics and contexts, and the responses in turn reflect the prevailing values and ideologies of society. This study focuses specifically on affect and emotion expression, rather than evaluative judgment, but still interprets the results as partly indicative of stance in political discourse and as a window to a wider societal picture. So stance is used in this study as an analytic lens that bridges individual expression with broader societal positionings.

Early psychological theoretical models of affect or emotion (e.g. Ekman 1992 and Izard 1977) provided only a limited set of basic emotions – typically anger, fear, disgust, sadness, happiness, and surprise – and associated each with distinct facial expressions and physiological responses. These approaches emphasized universality and biological grounding. Dimensional models (e.g., Russell 1980; Russell & Barrett 1999) conceptualized affect along continuous axes such as valence (pleasant–unpleasant) and arousal (high–low intensity).

Plutchik's psychoevolutionary model of emotion (Plutchik 1980, 1984, 2001) organizes eight primary emotions – joy, trust, fear, surprise, sadness, disgust, anger, and anticipation – into a circular structure known as the Wheel of Emotions, which represents their bipolar opposites (e.g., joy–sadness, anger–fear) and varying intensities (e.g., serenity–joy–ecstasy). It is one of the most influential models for categorizing and visualizing human emotions as it forms a basis from which any complex emotional states can be derived. The model has proven particularly useful in computational and corpus-linguistic research, where it provides a psychologically interpretable framework for classifying and aggregating emotion categories. Plutchik's wheel was used here as the basis for the emotion categories of the classifier model.

### `Earlier research`

Much of the earlier research on stance, affect, emotion, sentiment, or related concepts in parliaments has remained within the constraints of the positive-neutral-negative division while sometimes studying some additional variable of interest. This body of work also often highlights the well-established notion of *negativity bias*: the tendency for negative information to be more salient, attention-grabbing, and influential than positive information in human cognition and communication (Rozin & Royzman 2001). In political contexts, this bias often manifests as a disproportionate focus on problems, criticism, and conflict, making negative sentiment more prevalent in many forms of political text (Soroka 2014). Also, parliamentary emotion expression has been noted to increase during times when the MPs aim their speech more towards the broader electorate than just their fellow legislators (Osnabrügge *et al*. 2021).

Since the British Hansard record has been available for a long time and it is in English – the current *lingua franca* of the academic world – it has been studied much more than any other parliamentary corpus despite it being heavily edited. Rheault *et al.* (2016) measured emotional polarity and response to economic recessions in the Hansard and found that "the mood of politicians has become more positive during the past decades, and that variations in emotional polarity can be predicted by the state of the national economy" (ibid.: 1).

As records of other parliaments have become increasingly available as speech corpora, the variation of studied countries and languages has increased. The European Parliament has become a popular target of research due to its multilingual nature. Emotion in the European Parliament has been studied by Sanchez Saldago (2021), with results that suggest that the speaker's power position affects emotion expression, and that positive emotions were more typically expressed than negative emotions. The ParlaMint corpora (Erjavec *et al.* 2023, Erjavec *et al.* 2025) at present contain parliamentary speeches from 29 countries and autonomous regions in Europe. Eskişar and Çöltekin (2022) have used the data for studying emotions in the Turkish parliament, and they found that anger was the most dominant emotion and that "one can expect more emotional display by opposition parties, as doing so may help increase their credentials as a political opponent for potential supporters" (ibid.: 66). A literature review of earlier computational sentiment and position-taking analysis on parliamentary datasets has been presented by Abercrombie and Batista-Navarro (2020).

Few works have studied emotions in Finnish parliamentary data, which has been available for a few years in machine-readable format online (Sinikallio *et al.* 2023, regarding the creation of the infrastructure see Hyvönen *et al.* 2024). Tarkka *et al.* (2024) trained a model for emotion analysis on Finnish parliamentary debates, but the paper concentrated on the effects of model training methods rather than on the emotions themselves. A lexicon-based study on sentiment in the Finnish parliament by Lehtosalo and Nerbonne (2024) found that positive sentiment has been more common throughout the existence of *Eduskunta* (1907-2023 in Lehtosalo and Nerbonne's publication), and that positive sentiment has also increased over time. A study by Elo *et al.* (2025) studied two topics, economy and environmental politics, in *Eduskunta* speeches with a sentiment recognition deep learning model similar to one used in this paper and with the same emotion categories. They found several interesting patterns; firstly, that opposition MPs spoke with more negativity than government MPs, secondly, that both far-left and far-right MPs used more emotional speech, and thirdly that the categories hopeful-optimistic-trust and fear-worry-distrust dominated within the two examined topics.

To our knowledge, a full spectrum of topics has not been the main focus in any research regarding emotions in parliamentary discussions. However, studies on specific topics exist that could shed some light on what kind of emotions to expect in certain policy areas. Cheng *et al.* (2024) examined stance in legislative data protection laws in the United States, the European Union, and China. The results showed that stance choices tend to mirror the respective legal cultures, values, and ideologies in each country, but also that common to all was "a legislative tendency to achieve an overtly neutral appearance through covert stance expressions" (*ibid*: 1). Widmann (2025) studied German MPs' emotions in discussions regarding construction of wind turbines – a renewable energy source – and found strong emotional polarity between the strongest proponents and opponents. These studies suggest possible topic-specific emotion profiles: neutrality in legislation and polarity in energy discussions.

Regarding the underlying societal values that may reflect on parliamentary debates, recent studies and reports have indicated a shift toward more conservative and strict values in Finland and other Nordic countries. This trend is perhaps most visible in the emergence of populist radical right-wing political parties and their persistence in the field (e.g. Widfeldt 2023; Widfeldt 2018; Jungar & Jupskås 2014). Also, neoliberal values – the belief in the superior efficiency and legitimacy of markets, individual responsibility, competition etc. – seem to have become mainstream among the Finnish conservative parties since their "breakthrough" in the 1990s (Kärrylä 2024), and the generation of Millennials (born in the 1980s and the 1990s) have been shown to have adopted these values along with some nationalistic tendencies (Helve 2023). A report on the 2023 Finnish parliamentary election suggested that a "right turn" took place in the voting behavior (Grönlund & Strandberg 2023), and an analysis on the voters (Strandberg 2023) indicated clear affective polarization between voters representing value-liberalism and value-conservatism.

Despite the developments of increasing conservative values and polarization, Lehtosalo and Nerbonne's (2024) findings indicated that positive sentiment has increased over time. Similar results were reported by Rheault *et al.* (2016) regarding the parliamentary debates in the UK between 1909 and 2013. On the other hand, Evkoski *et al.* (2025) have reported increasing negativity in the parliaments of former Yugoslavian countries. Diachronic changes in specific emotions have not been studied in parliaments as far as we know.

**Topic landscape**

Topic selection, or what is being talked about, is not solely an individual speaker's choice because every parliamentary plenary session has a pre-defined agenda. It is constructed by the Speaker's Council (*Puhemiesneuvosto*) and specifies in advance which bills, interpellations, government reports, or other items need to be discussed. So, while the macro-level topic selection is institutionally determined and while individual speakers can mix themes from other topics into the discussion, the MPs have the most agency over *how* they talk about the topics at hand.

A politician can be expected to use emotion expression deliberately. It is possible that the topics and contexts under discussion evoke emotions and sometimes the speaker simply expresses them as they come, but in the parliamentary context emotion expression is typically used as a tool to also produce emotional responses in others, for example. We can also assume that MPs will likely invest emotion expression in topics they consider important or strategically relevant (Sperber & Wilson 1995). Mapping emotions across topics reveals such "hot topics" in the parliamentary topic landscape.

A 'topic landscape' as a concept (earlier used in Ristilä 2025) refers to the overall distribution of topics within a corpus, enriched by one or more layers of variable correlation. Rather than focusing on the technical procedures of topic extraction, the topic landscape perspective foregrounds the mapping of how topics interact with contextual, temporal, or speaker-related factors. A topic landscape analysis may be either univariate, in which a single variable of interest (e.g. emotion) is examined across all topics, or multivariate, in which multiple dimensions jointly shape the landscape and possibly reveal more detailed links and groupings between variables. This allows both analyses that explain a particular variable through topics ("the topic landscape of X") and broader descriptions of the corpus as a multidimensional thematic terrain.

In this study the focus is on the topic landscape of emotions. This perspective enables an examination of how emotions distribute across policy areas, revealing affective patterns that may be related to deeper societal values.

# Data and Methodology

The previous section introduced earlier research, the terminology related to emotion analysis in general and more specifically the key concepts used in this study. This section describes the key features of the data used in this study and explains the methods chosen to conduct the analyses.

The data consisted of Finnish parliamentary plenary speeches from years 2000 to 2020 (Sinikallio *et al*. 2023) with metadata, including date of speech and speaker info. The years were selected after an initial modeling test which revealed that the emotion model got confused by older data, probably due to differences in expressions and vocabulary.

To ensure the analysis centered on substantive parliamentary debate rather than procedural content, speeches by the Speaker of Parliament (*puhemies*) and speeches with fewer than five sentences were excluded from the corpus (79,050 speeches filtered out in total). This resulted in 218,857 speeches and 3,761,491 sentences. This filtered corpus is used in all the analyses.

Using digital tools was necessary, since the corpus would have been too large to be analyzed by one or even multiple human annotators. This study combined results from two digital modeling approaches: emotion modeling and topic modeling. Emotion modeling can be understood as a more fine-grained form of sentiment analysis; whereas sentiment analysis typically distinguishes between positive, negative, and neutral valence, emotion modeling identifies more specific emotional categories, such as 'hope', or 'fear', allowing for a more detailed examination of affective expression. Topic modeling is a computational method used to identify latent thematic structures in large text collections by grouping words that tend to co-occur across documents. It operates on the assumption that each text can be represented as a mixture of multiple topics, and that each topic consists of a set of words with varying probabilities. Topic modeling allows researchers to explore the conceptual organization of a corpus without predefined categories.

The emotion model used in this study was a custom model specifically fine-tuned for the corpus. It used XML-R-parla (Mochtak *et al*. 2023) as its base model and was further fine-tuned with 3,415 sentences from the Finnish parliamentary speeches given between years 2010 and 2020 and annotated by human raters into nine categories. The same categories were used by the model as category labels when analyzing new data:

four positive

- joy-success (JOY-)
- hopeful-optimistic-trust (HOPE)
- love-compliments (LOVE)
- positive_surprise (POSI)

four negative

- sadness-disappointment (SADN)
- fear-worry-distrust (FEAR)
- hate-disgust-taunts-mockery (HATE)
- negative_surprise (NEGA)

and one neutral (NEUT) category. The model achieved a micro F1 score of 0.63, representing a moderate but adequate level of performance. While emotion models provide scalable and systematic ways of capturing emotion, they do not "understand" language in the human sense and therefore often fail with more subtle or context-dependent phenomena such as irony, sarcasm, or rhetorical understatement. As a result, their classifications may at times oversimplify or even misrepresent the emotion that is being conveyed. Critical awareness is therefore essential when interpreting computationally derived results.

The topic modeling method used in this study was Latent Dirichlet Allocation (LDA; Blei, Ng & Jordan 2003). The model produced 26 topics, which are listed in **Appendix A** along with the most important words belonging to each topic. More details of the creation of the topic model that was used here are presented in (Ristilä & Elo 2023). Topics were defined for larger collections of sentences, while emotions were initially defined per sentence. This is because the used topic modeling method works better with longer texts, while emotion models work better with short texts.

A small portion of text (<1%) was in Swedish, as the Finnish parliament is bilingual. The emotion model could process both Finnish and Swedish, but the topic model was monolingual in Finnish and could not understand Swedish words, but because there was so little Swedish this was not seen as an issue.

The analysis was conducted from two perspectives, synchronic (treating the full corpus as one point in time) and diachronic (observing changes over time). The analysis steps for each perspective are described below.

For the synchronic analysis, first the basic emotion category distribution of the full filtered corpus was calculated by simply dividing the number of sentences belonging to each category with the total number of sentences. Second, for the synchronic topic-emotion analysis, sentences were grouped into emotion subcorpora based on their predicted labels, and each emotion subcorpus was then processed with the topic model, resulting in a topic distribution for each emotion category.

By creating a cross-table of the topic distributions across emotion subcorpora and comparing each topic-emotion prevalence to the average topic prevalences, it was possible to identify topics that were over- or underrepresented in some emotions. To obtain average prevalences for each topic that were comparable across emotion subcorpora, the prevalences of a topic in each emotion subcorpora were summed and divided by the number of emotion categories. Simple differences from these averages were converted into percentages for better comparison. Topics were then grouped into five groups based on their overrepresentation: *negatively skewed*, *neutrally skewed*, *positively skewed*, *polarized*, and *average*.

Due to the extremely large sample sizes in the emotion corpora (ranging from tens of thousands to nearly a million sentences), even small deviation from the expected (average) proportions would have resulted in extremely large z-scores and effectively zero p-values. Therefore, traditional statistical significance testing was not informative in this context. Simply reporting the observed ratios provided a more interpretable and practically relevant measure of topic over- and underrepresentation.

A single emotion category was considered skewed if there was at least 50% over- or underrepresentation, a threshold deemed large enough to indicate substantive deviation. Negative skewness was determined if a topic was at least 50% overrepresented in one or more negative emotion categories but not in positive or neutral categories. Similarly, if overrepresentation emerged in positive categories but not in negative or neutral, the topic was considered positively skewed. Topics showing only neutral overrepresentation were considered neutrally skewed. If a topic was overrepresented in neutral and either in a positive or negative category, the grouping was assigned based on the highest overrepresentation. If overrepresentation was observed in both positive and negative categories but not neural, the topic was considered polarized. If no overrepresentation (50% or higher) in any category was observed, the topic was considered average. Underrepresentation was considered a secondary supporting factor when interpreting the groupings.

For the diachronic analysis, first an average weighted sentiment (WAS) score was calculated so that the general positive–negative tone of the parliamentary discussion could be examined over time. Using an average sentiment meant temporarily losing emotion-specific information but gaining the ability to compare results with earlier research, where the positive–negative level has been dominant. The WAS score was calculated for each speech and then mapped over time. The method to calculate WAS is presented in **Formula 1**.

$$WAS = \frac{\sum_{i=1}^{n}(S_{sent} * S_{prob})}{n_S}$$

- **S**$_{sent}$ denotes a sentiment multiplier; 1 for a sentence with positive emotion label, -1 for a negative label, and 0 for a neutral label.
- **S**$_{prob}$ denotes the probability, or value of certainty, of the predicted label; values range between 0 and 1.
- **n**$_s$ denotes the number of sentences in a speech.

**Formula1**: *Weighted average sentiment (WAS) of a speech.*

Second, separate emotion category levels over time were calculated and graphed as monthly three-month averages by summing emotion category probabilities within the three-month window, then dividing that sum by the sum of all emotion category probabilities within the same time window. This reduced short-term noise but still reflected the broader patterns of parliamentary activity. Since debates are often clustered around legislative cycles and election periods, light averaging helps capture the broader dynamics of parliamentary discourse (Korhonen, Kotze & Tyrkkö 2023:6).

Finally, the changes in topic proportions within each emotion category were graphed to see which topics gained or lost expression in emotion categories over time. These graphs were also calculated as monthly three-month averages. All sentences representing an emotion category form a three-month window were combined into a "document" that was processed with the topic model, producing a topic distribution. Each monthly window produced a topic distribution, and from this series of distributions each topic could be separated into its own topic-emotion graph.

Close reading was used to support the analyses because the emotion model used was relatively opaque compared to a lexicon and did not inherently explain its classifications. Close reading was used as a type of verification step to determine whether the emotions suggested by the model could also be identified qualitatively around the topics implied.

# Results

This section describes the findings on how emotions are patterned across the Finnish parliamentary topic landscape. First, the synchronic emotion distribution overall and the topic-emotion distributions are given, second the observed diachronic changes are presented in emotions alone as well as in the topic-emotion relationships.

## Synchronic landscape

Regarding emotion expression in the corpus overall, certain categories were significantly more common than others, as seen from **Figure 1**. Most of the expression was *neutral* (NEUT, ca. 24%) despite filtering out the Speaker of Parliament speeches and short speeches. The second most expressed emotion category was *hopeful-optimistic-trust* (HOPE, ca. 23%), the third most expressed category was *fear-worry-distrust* (FEAR, ca. 21%), followed by *hate-disgust-taunts-mockery* (HATE, ca. 14%), while the rest were all under 10% each. *Positive_surprise* (POSI) expression rate was only 0.002%. It is notable that out of all the positive emotion categories HOPE was substantially more expressed than the others, as all other positive category expression rates were under 5%. Examples of HOPE in various contexts are presented later in examples 1, 3, 6, 8, 10, 11 and 16.

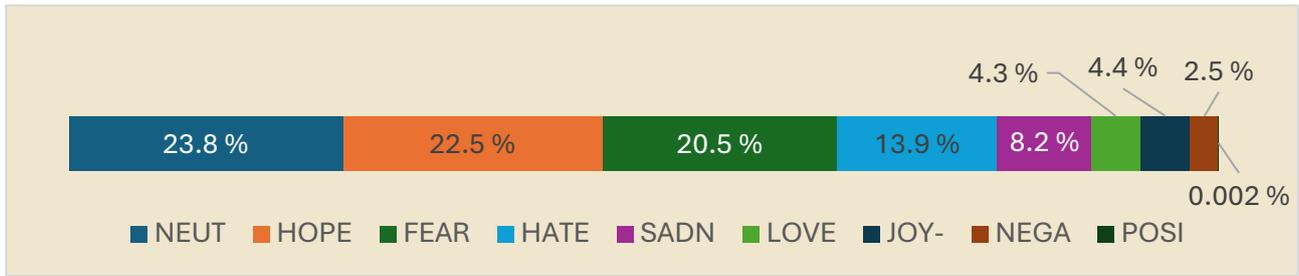

*Figure 1*: Distribution of emotion categories in a corpus where Speaker of Parliament speeches and speeches with fewer than 5 sentences had been removed.

Aggregating all positive and all negative emotions provided a broad sentiment overview for comparison with earlier work, as sentiment has long been the dominant approach in computational studies of political discourse. The proportions – 31% positive, 45% negative, and 24% neutral – showed that negativity, or negative stance, was expressed more frequently than positive stance, and that emotional expression overall was more common than neutrality. This pattern is consistent with some of the prior findings that parliamentary and legislative discourse tend to display negativity bias, reflecting the adversarial, problem-oriented nature of political debate (Soroka 2014).

As emotion expression was expected to be partly explainable by topics, the topic-emotion distributions were examined. Simple differences in how much more or less each topic appeared in each emotion category relative to the overall distribution were calculated, then converted into percentages for better comparison. These proportional changes are presented in **Table 1**. The raw prevalences are presented in **Appendix B**. Magnitudes of the differences ranged from -0.98 to 3.44 (or from -98% to 344%), indicating significant variation between topic representations within emotion categories. Interpreting this from the topic perspective suggests that some policy areas attract systematically distinct emotional expressions.

Aggregating the topics based on their overrepresentation revealed five loose sentiment skewness groups: *polarized*, *negatively skewed*, *neutrally skewed*, *positively skewed*, and *average*, also given in **Table 1**. If a topic was represented in an emotion category over 50% more than average, it counted towards skewness in that sentiment grouping, and if both positive and negative categories showed skewness, the topic was considered polarized.

*Table 1.* Heat map of proportional differences of topic prevalences in the subcorpora from overall topic averages. Differences over +0.5 have been bolded. The raw topic prevalence table on which this is based on is presented in Appendix B.

|  | topic | HATE | SADN | FEAR | NEGA | NEUT | LOVE | HOPE | JOY- | POSI |
|---|---|---|---|---|---|---|---|---|---|---|
| polarized | Employment | 0.00 | 0.46 | **0.62** | -0.24 | -0.13 | -0.40 | **0.60** | 0.07 | -0.98 |
|  | Energy | -0.11 | -0.32 | **0.87** | -0.34 | -0.23 | -0.45 | **1.20** | -0.27 | -0.34 |
| negatively skewed | Parliamentary factions | **0.54** | 0.28 | -0.33 | **1.02** | -0.63 | 0.02 | -0.51 | -0.15 | -0.24 |
|  | Budget | -0.13 | 0.28 | **0.98** | -0.38 | -0.06 | -0.58 | 0.18 | 0.30 | -0.58 |
|  | Taxation | **0.59** | 0.33 | 0.40 | -0.10 | 0.33 | -0.59 | -0.18 | -0.28 | -0.52 |
| neutrally skewed | Crime | 0.05 | 0.50 | 0.15 | -0.26 | **0.72** | -0.09 | -0.38 | -0.21 | -0.48 |
|  | Law proposals | 0.06 | -0.48 | -0.55 | -0.01 | **3.44** | -0.40 | -0.64 | -0.45 | -0.96 |
|  | Legislation | 0.00 | -0.05 | -0.07 | -0.13 | **1.36** | -0.12 | -0.33 | -0.32 | -0.34 |
|  | Traffic and transport | -0.29 | 0.11 | -0.03 | -0.45 | **1.11** | -0.19 | 0.15 | 0.14 | -0.56 |
|  | Social and health care | 0.06 | 0.11 | 0.08 | -0.29 | **0.80** | -0.14 | 0.07 | -0.15 | -0.53 |
|  | Question time | -0.14 | -0.29 | -0.47 | 0.06 | **1.06** | **0.81** | -0.60 | 0.05 | -0.48 |
|  | Pensions | -0.13 | 0.05 | 0.06 | -0.40 | **0.95** | **0.54** | -0.38 | -0.16 | -0.51 |
| positively skewed | Foreign and security policy | -0.26 | -0.29 | -0.04 | -0.37 | 0.15 | 0.08 | **0.54** | 0.38 | -0.19 |
|  | Public sector | -0.35 | -0.02 | 0.31 | -0.45 | -0.03 | -0.09 | **0.90** | 0.43 | -0.70 |
|  | Agriculture | -0.20 | 0.24 | 0.36 | -0.33 | -0.01 | -0.10 | **0.52** | 0.09 | -0.57 |
|  | Development cooperation | -0.34 | -0.31 | -0.37 | -0.46 | 0.00 | 0.39 | -0.05 | **0.63** | **0.51** |
| average | Social problems | -0.13 | -0.08 | -0.05 | -0.24 | -0.35 | -0.25 | -0.46 | -0.36 | **1.91** |
|  | Education | -0.41 | -0.14 | -0.30 | -0.47 | 0.02 | -0.10 | 0.23 | 0.21 | **0.96** |
|  | Housing | 0.15 | -0.01 | 0.16 | -0.28 | -0.27 | -0.50 | -0.21 | -0.18 | **1.14** |
|  | Regionality | -0.27 | 0.05 | 0.07 | -0.28 | -0.20 | -0.22 | 0.11 | 0.08 | **0.65** |
|  | GENERAL | 0.16 | -0.11 | -0.26 | 0.38 | -0.43 | 0.29 | -0.43 | -0.11 | **0.52** |
|  | Administration | -0.08 | -0.07 | -0.14 | -0.10 | 0.20 | -0.02 | -0.16 | -0.03 | 0.41 |
|  | Commerce | 0.22 | -0.04 | 0.37 | -0.43 | 0.16 | -0.58 | 0.38 | -0.13 | 0.05 |
|  | Democracy | 0.27 | -0.17 | -0.35 | 0.16 | 0.32 | 0.37 | -0.20 | 0.03 | -0.43 |
|  | Social benefits | -0.46 | 0.43 | 0.42 | -0.56 | -0.12 | -0.37 | 0.48 | -0.15 | 0.33 |
|  | Voting | -0.14 | -0.08 | -0.42 | 0.03 | 0.37 | 0.34 | -0.46 | -0.10 | 0.46 |

The *polarized* group contained only two clear cases: "employment" and "energy", both overrepresented in the two largest non-neutral categories, HOPE and FEAR. "Energy" was extremely polarized (+0.87 FEAR, +1.20 HOPE), reflecting how any energy solution can have far reaching consequences and multiple sides to it, as highlighted in the differing views on wood burning in **Example 1** and **Example 2**. "Energy" topic is also related to "employment", because energy production needs labor and creates new jobs. This connection is highlighted in **Example 3** where peat production's impact on employment is discussed.

> **Example 1** (*positively skewed*: HOPE, "energy"): *Puuenergian käytön lisääminen kaikissa muodoissaan on kannatettava asia, josta lienee täysi kansallinen yksimielisyys kaikissa poliittisissa ryhmissä ja kansalaispiireissä.* "Increasing the use of wood energy in all its forms is a worthy cause, one that will likely enjoy full national consensus across all political groups and civic circles."

> **Example 2** (negatively skewed: FEAR, "energy"): *Jos omakotitalon lämmittäjä ostaa halot ja klapit, hän joutuu maksamaan arvonlisäveron ja loppukäyttäjänä hän ei voi tätä ikävä kyllä vähentää mistään.* "If a homeowner buys logs and firewood, they have to pay VAT and, as the end user, unfortunately they cannot deduct this from anything."

> **Example 3** (positively skewed: HOPE, "employment" and "energy"): *Keski-Suomen kunnissa on turpeella merkittävä taloudellinen työllistävä vaikutus erityisesti luoteisen Keski-Suomen ja pohjoisen Keski-Suomen kunnissa, kuten Pylkönmäki, Kyyjärvi, Karstula aivan erityisesti ja eräät muut pitäjät.* "In the municipalities of Central Finland, peat has a significant economic impact on employment, especially in the municipalities of northwestern Central Finland and northern Central Finland, such as Pylkönmäki, Kyyjärvi, Karstula in particular, and some other municipalities."

The *negative skewness* group included topics "parliamentary factions", "budget" and "taxation", suggesting that speech related to other political parties or allocation of money typically elicits more negative emotions, highlighted in **Example 4** and **Example 5**. In Example 4 the speaker expresses worry that increasing taxes will worsen the economic situation of municipalities, and in Example 5 the speaker expresses negative surprise on the apparent lack of cooperation.

> **Example 4** (*negative skewness*: FEAR, "budget" and "taxation"): *Mutta luulisin, kun todellisuudessa nyt veroäyrin hinnat nostetaan 21—22 penniin taikka veroprosentti, niin kuin nykyisin virallisesti sanotaan, silloinhan se nähdään vasta, kuinka todella heikko kuntien tilanne on.* "But I think that when the tax rates are actually raised to 21-22 pennies, or percent, as they are now officially saying, then we will only see how weak the situation of the municipalities really is."

> **Example 5** (*negative skewness*: NEGA, "parliamentary factions"): *Kun on katsonut Suomen keskustan ja kokoomuksen kamppailua porvariyhteistyöstä, jotenkin vasemmistolaisena ihmisenä ja köyhempien edustajana ja kuitenkin inhimillisen ajattelun ihmisenä käy ihmettelemään, onko tosiaan menossa näinpäin, että punamultayhteistyö on haudattu kokonaan.* "After watching the struggle between the Finnish Center Party and the National Coalition Party over bourgeois cooperation, somehow as a leftist person and a representative of the poor, and yet as a person of humane thinking, one wonders whether it is really going in this direction that 'red soil cooperation' has been completely buried."

"Crime" also showed negative skewness because it was overrepresented in SADN (+0.5), but because its overrepresentation in NEUT was stronger (+0.72) it was grouped into *neutral skewness*. Based on close reading, neutrality in the "crime" topic may have emerged because multiple law proposals affecting the Criminal Code were processed during the observed period, as exemplified in **Example 6** where a suggested addition to the penal code is stated matter-of-factly. "Law proposals" was a strongly neutrally skewed topic, showing the highest overrepresentation in the corpus (NEUT +3.44). Other topics in the neutral skewness group included "legislation" and "question time", that together with "law proposals" represent topics that contain abundant procedural speech. Unexpectedly, much less procedural topics "traffic and transport", "social and health care", and "pensions" were also neutrally skewed. It is not possible to say for certain with only cursory close reading, but this may be partly due to related law proposals or perhaps because discussing these topics simply does not elicit as much emotion expression as others. "Question time" showed high LOVE which seemed to stem from politeness as speakers often thanked other commenters: the phrase "thank you for the question" (*kiitos kysymyksestä*) alone appeared 25 times in the LOVE corpus.

> **Example 6** (*neutral skewness*: NEUT, "crime"): *Rikoslakiin ehdotetaan lisättäväksi, että aikuisille tarkoitetun kuvaohjelman esittäminen tai levittäminen lapsille tulee rangaistavaksi.* "An addition to the Criminal Code is proposed: that showing or distributing to children visual content that is intended for adults is to become punishable."

The *positive skewness* group was aggregated based on all other positive emotion categories except POSI, as the POSI category contained so few sentences (77) that differences in its topic distribution were not significant or in any way comparable with the other categories. Topics with clear positive skewness were "foreign and security policy", "public sector", "agriculture" and "development cooperation"; only "development cooperation" showed overrepresentation in JOY-, while the others overrepresented HOPE, the second largest emotion category. "Development cooperation" and "foreign and security policy" also showed notable underrepresentation in all negative categories, suggesting that discussions regarding matters outside the domestic sphere may show the most positivity. This view is supported by **Example 7** and **Example 8**, where joy of success is expressed about Germany increasing development aid (**Example 7**) and strong optimism is expressed over northern railway connections (**Example 8**).

> **Example 7** (*positive skewness*: JOY-, "development cooperation"): *Ainakin Saksassa tämä suositus on otettu todesta ja jopa Saksan pääministeri on ottanut julkisesti kantaa kehitysavun nopean nostamisen puolesta.* "At least in Germany, this recommendation has been taken seriously and even the German Prime Minister has taken a public stand in favor of a rapid increase in development aid."

> **Example 8** (*positive skewness*: HOPE, "foreign and security policy"): *Norjalaiset ovat myös ilmaisseet kiinnostuksensa rautatieyhteyksien rakentamiseen, niin että Kolariin päättyvää rataa jatkettaisiin aina Jäämerelle asti Storfjordiin. (…) Tällaiset näkökulmat avautuvat pohjoisesta. Ne ovat valtavien mahdollisuuksien näköalat, ja jotakin noista mahdollisuuksista päästään toteuttamaan, jos pidetään huoli siitä, että suhteet naapureihin säilyvät hyvinä ja, voisiko sanoa, uutta luovina.* "The Norwegians have also expressed interest in building railway connections, so that the line ending in Kolari would be extended all the way to the Arctic Ocean at Storfjord. (…) Such vistas open up from the north. They are vistas of enormous possibilities, and some of those possibilities can be realized if care is taken to ensure that relations with the neighbors remain good and, one might say, innovative."

Topics that showed no overrepresentation in any emotion category were "administration", "commerce", "democracy", "social benefits", and "voting". "Commerce" and "social benefits" showed hints of polarization, as weak overrepresentation was observed in HOPE and FEAR. On the other hand, "voting" was slightly underrepresented in HOPE and FEAR. For clarity, topics overrepresented only in the POSI category, namely "social problems", "regionality", "education", "housing", and "general", were grouped as a subgroup of *average* and marked in light red in **Table 1**. Though POSI was extremely rare, it seemed to surface in anecdotal segments of speech, such as in **Example 9** where an MP reminisces their early years as a university student.

> **Example 9** (*average*/POSI, "education"): *Kun 50-luvulla aloitin opiskelun yliopistossa, silloin koin erittäin voimakkaana sen, kuinka valtavan upea tämä maksuttomuus on, kuinka köyhästä perheestä ja varakkaasta perheestä tuleva lahjakas, sanotaanko riittävän lahjakas, ihminen pystyi opiskelemaan ja suorittamaan tutkintonsa.* "When I started studying at university in the 50s, I experienced very strongly how incredibly wonderful this free education is, how a talented, shall we say sufficiently talented, person from a poor family and a wealthy family was able to study and complete their degree."

This section presented the findings of the synchronic perspective, namely the emotion category distribution and the five groups of topics based on emotion over- or underrepresentation. The next section adds the perspective of time and examines the emotion categories and topic-emotion pairs from the diachronic perspective.

## Diachronic changes in the landscape

In this section the diachronic findings will be presented. Firstly, the weighted average sentiment (WAS) over time is given, secondly expression rates of the different emotion categories over time are presented, and lastly changes in the topic prevalences within the emotion categories (topic-emotion pairs) are introduced.

### *Average sentiment and average emotion categories over time*

A three-month moving average of the weighted average sentiment (WAS) over time, presented in black in **Figure 2**, showed that despite remaining below zero (negative) for most of the observed period, parliamentary speech has become less negative. The development towards positivity partly aligns with a similar result by Lehtosalo and Nerbonne (2024) and suggests a gradual shift toward a more optimistic stance despite the overall negativity. The three-month average also showed substantial variation, most notably a sharp decrease around the 2011 election. The 2011 election result was special as it was the year when the right-wing party True Finns (*Perussuomalaiset*) got a landslide of votes and got into the government for the first time. The yearly average showed a clear dip in 2015, which was not an election year but may be related to the European Refugee Crisis. Interestingly, the year 2020, when Covid affected Finland the most, showed only a moderate decline in both the three-month WAS as well as the yearly WAS.

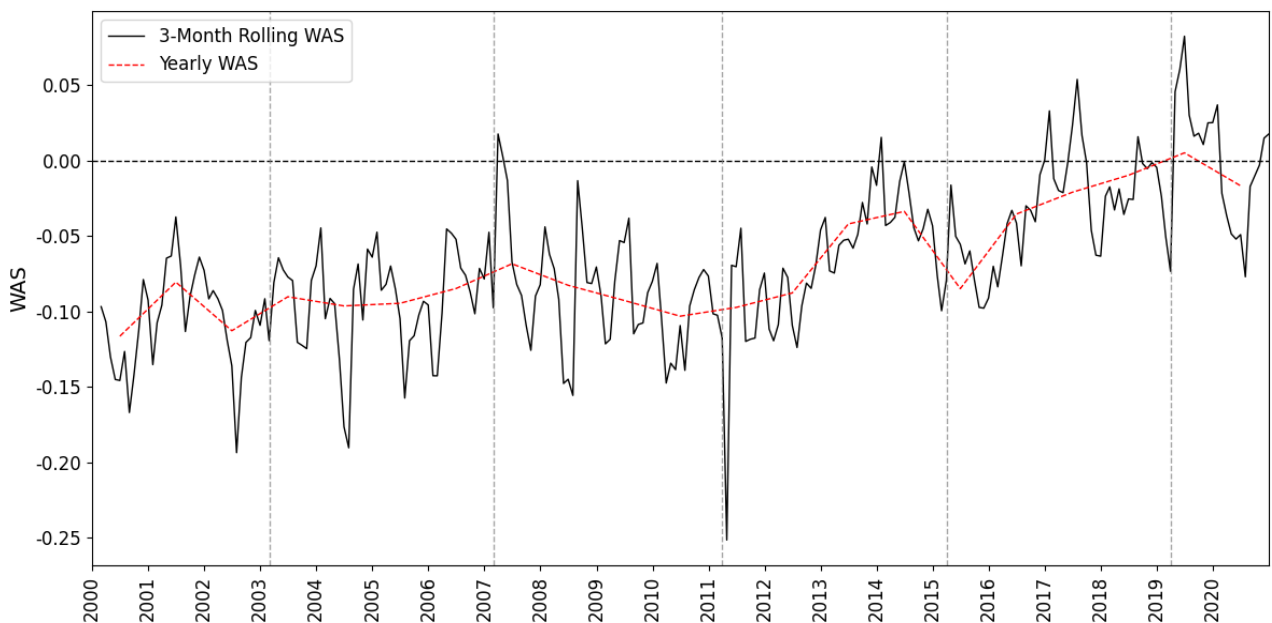

*Figure 2*. Monthly moving 3-month weighted average sentiment (WAS) in black, yearly WAS in red. Election dates have been marked with dashed vertical lines.

Expression rates of the different emotion categories (**Figure 3**) illustrate the constituent dynamics of the average sentiment. The most notable developments emerged in the HOPE category (hopeful-optimistic-trust). First, HOPE displayed a relatively steady increase across the observed period, explaining much of the increase in the overall WAS. Second, the category showed fluctuations around election years (2003, 2007, 2011, 2015, and 2019): in each case, HOPE declined somewhat preceding the election, rose sharply around the election itself, and then shortly afterwards returned close to what it was before. To reveal the factors behind this phenomenon would require detailed close reading of the themes and contexts of the discussion around each election, which is outside the scope of this paper. Also, the fact that *Eduskunta* does not convene for 3-4 weeks before elections may have had an effect as well. The post-election increases in HOPE may reflect optimism associated with a change of government, as illustrated in **Example 10**, as well as the surge of promises and forward-looking commitments typically presented in newly announced government programs, as shown in **Example 11**.

> **Example 10** (HOPE, "social problems"): *Toivon kuitenkin, että kun tämä uusi hallitus näyttää jo mainostavan itseään uuden käyttäytymiskulttuurin hallituksena, että sen merkittävät ministerit ja poliitikot näyttäisivät esimerkkiä ja pitäisivät linjapuheensa tulevaisuudessa eduskunnassa.* "However, I hope that as this new government already seems to be promoting itself as a government of a new behavioral culture, that its prominent ministers and politicians will set an example and deliver their policy speeches in Parliament in the future."

> **Example 11** (HOPE, "social problems"): *Olemme asettaneet kunnianhimoiset tavoitteet uuden hallituksen työlle. (…) Hallituksen tavoitteena on vastuullinen, välittävä ja kannustava Suomi. (…) Uskomme, että parempia tuloksia saadaan aikaan käyttämällä positiivisia kannustimia*. "We have set ambitious goals for the work of the new government. (…) The government's goal is a responsible, caring and encouraging Finland. (…) We believe that better results can be achieved by using positive incentives."

Two other emotion categories also seemed to fluctuate around elections. These patterns are noted here, but, like with the case of HOPE, it is not possible to make any certain claims regarding the causes behind them. FEAR declined before elections and usually rose gradually back after a short period that typically matched the time HOPE took to come back down. Year 2011 was an exception, as FEAR jumped up immediately after the election. NEUT also seemed to inversely follow HOPE around elections, as NEUT increased right before elections and dipped right after them. Although the causal mechanisms cannot be directly inferred, these three large categories – NEUT, HOPE, and FEAR – may be somewhat linked to each other.

Other emotion categories showed more long-term changes. The period before mid-2002 differed from the rest of the observed period as FEAR was higher and HATE was lower than in other times. FEAR decreased in the period of 2011-2019 but HATE only decreased between 2011 and 2015. Instead, LOVE seems to have increased during the same period. The year 2020 showed marked fluctuations – HOPE decreased, FEAR increased, and HATE decreased – probably in the wake of Covid-19.

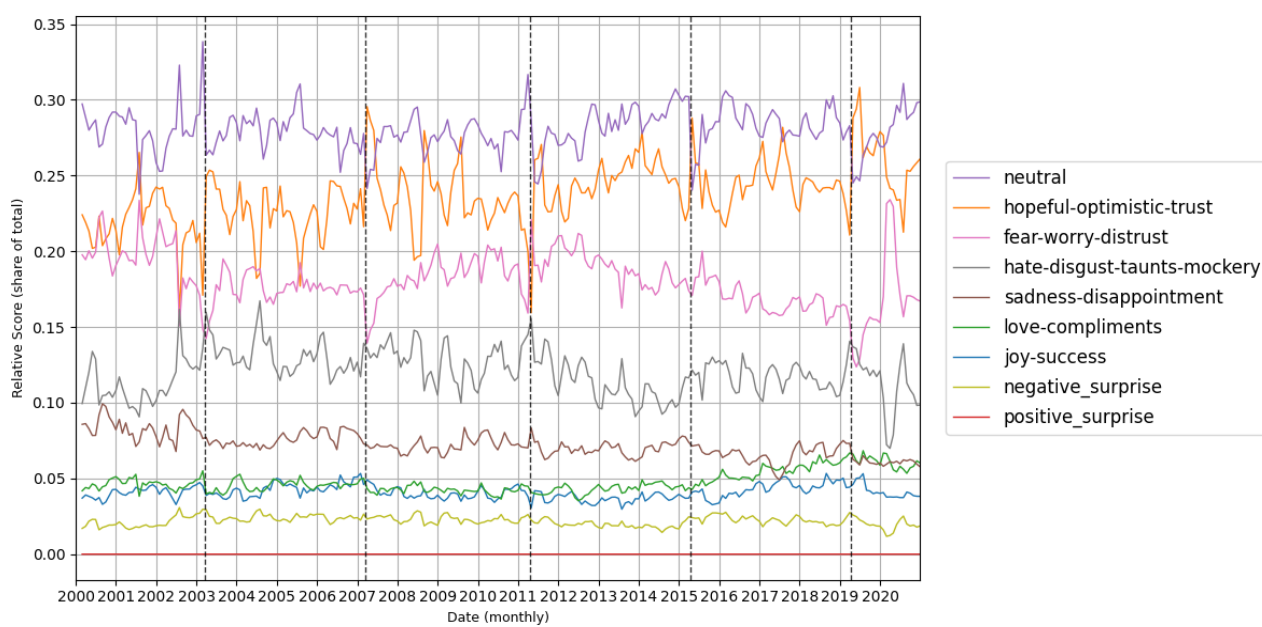

*Figure 3*. Monthly 3-month averages of weighted average emotion expressions. Vertical dashed lines mark election dates.

`Topic-emotion prevalences`

Lastly – to observe what has changed in the detailed topic landscape – topic prevalences over time within the emotion subcorpora were examined. Graphs for all topic-emotion pairings are provided in **Appendix C**; only findings that showed meaningful temporal patterns are discussed in the main text. A topic-emotion trend was considered meaningful if a regression line fitted to the monthly three-month window prevalences showed at least moderate explanatory power ($R^2 > 0.3$) and if the corresponding regression coefficient was statistically significant ($p < 0.05$); only few topic-emotion pairings filled these requirements.

The "parliamentary factions" topic showed meaningful decline in all negative emotion categories – FEAR, HATE, SADN, and NEGA – but also in JOY- (**Figures 4a-e**). This means that though "parliamentary factions" was inherently negative in tone as a whole, general negativity towards opposing parties may have been decreasing over the observed period. Close reading provided hints of decreasing sharpness of words also within categories, exemplified by **Examples 12**, **13**, and **14**: all examples are from the HATE category and in each one the Center Party is being criticized for following the National Coalition Party's suit, but the tone becomes increasingly neutral in the later examples.

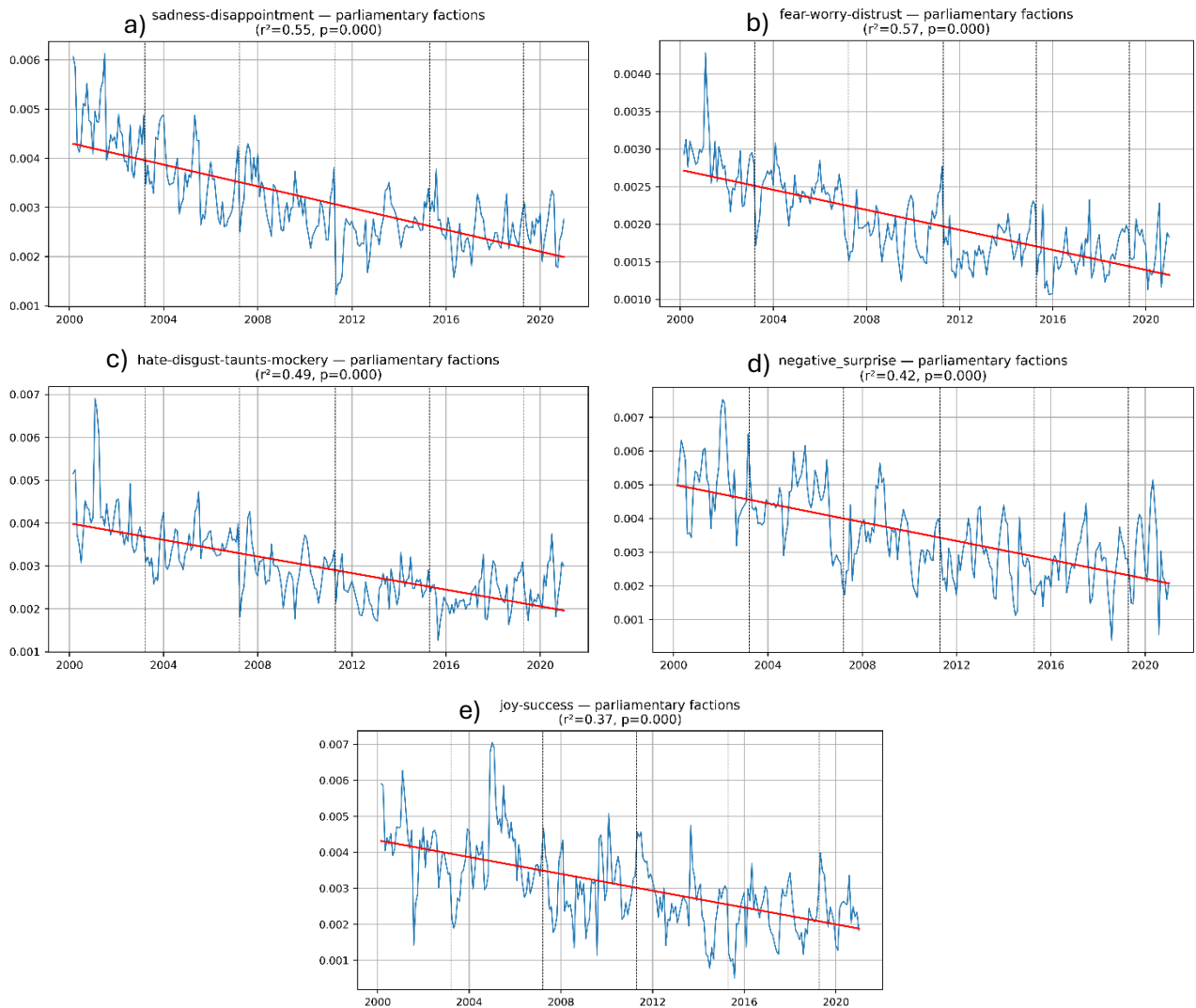

*Figure 4, a–e.* Topic-emotion graphs for topic "parliamentary factions" in emotion categories SADN, FEAR, HATE, NEGA, and JOY-.

**Example 12** (HATE, "parliamentary factions", 2000-03-22): *Ei ole montakaan viikkoa siitä, kun keskusta flirttaili kokoomuksen kanssa — tunne taisi olla molemminpuolinen — ja sitä oltiin niin porvaria, että vasemmalla oikein hirvitti.* "It hasn't been many weeks since the Center Party was flirting with the National Coalition Party — the feeling seems to be mutual — and it was so bourgeois that the left was actually frightened."

**Example 13** (HATE, "parliamentary factions", 2009-06-16): *Toivon, että Keskusta ryhdistäytyisi ja lopettaisi Kokoomuksen myötäilyn (…).* "I wish that the Center Party would get a grip of themselves and stop supporting the National Coalition Party (…)."

**Example 14** (HATE, "parliamentary factions", 2019-12-17) *Keskusta oli vielä eilen kokoomuksen kanssa samaa mieltä.* "The Center Party was still in agreement with the National Coalition Party yesterday."

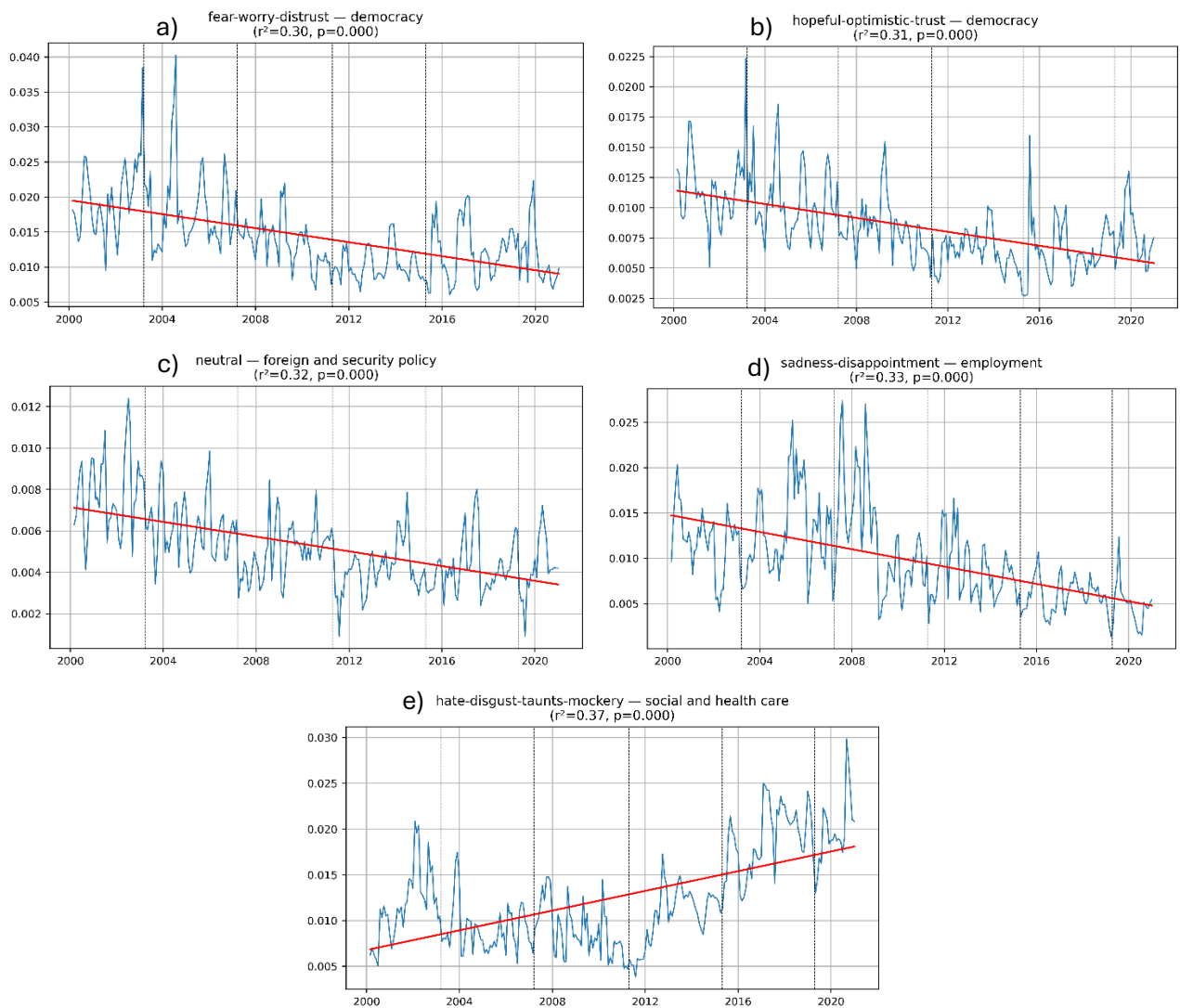

*Figure 5, a-e*. Topic-emotion graphs for a) FEAR—"democracy", b) HOPE—"democracy", c) NEUT—"foreign and security policy", d) SADN—"employment", and e) HATE—"social and health care".

Five other topic-emotion pairs also showed declining trends. The topic "democracy" showed decreasing trends in both HOPE and FEAR (**Figures 5a** and **5b**). As these emotion categories are of the opposing sentiment groups, this development suggests decreasing polarization of the topic. **Examples 15** and **16** show how "democracy" is a versatile topic and it can be used to evaluate different issues, such as political trends or points in a government program. If democracy is used less frequently as an evaluative frame, this may suggest that debate has shifted from value-oriented toward more efficiency-oriented frames, potentially reflecting a broader hardening of political values.

> **Example 15** (FEAR, "democracy", 2000-03-07): *Vapauspuolueen edustama äärioikeistolainen populismi herättää vakavia kysymyksiä demokratian rajoista.* "The far-right populism represented by the Freedom Party raises serious questions about the limits of democracy."
>
> **Example 16** (HOPE, "democracy", 2007-04-25): *Hallitusohjelmaan kirjattu alueiden kehittämispoliittisen toimivallan siirtäminen kansanvaltaiseen ohjaukseen on tervetullut muutos lähidemokratian lisäämiseksi ja valtion byrokratian purkamiseksi.* "The transfer of regional development policy authority to democratic control, as included in the government program, is a welcome change to increase local democracy and dismantle state bureaucracy."

Topic "foreign and security policy" showed a declining trend within NEUT (**Figure 5c**), meaning that the topic is spoken less in a neutral tone. Since the topic as a whole showed overrepresentation in HOPE, it is possible that the neutrality has shifted towards hopefulness. Close reading somewhat supports this, as sentences labelled as NEUT and HOPE are often quite similar, as shown in **Example 17** and **Example 18**. Both examples are about cooperation between Russia and one or more Nordic country, and neither expresses emotion very directly.

> **Example 17** (NEUT, "foreign and security policy", 2012-06-12): *Norja ja Venäjä ovat juuri sopineet yhteisestä viisumittomasta raja-alueesta molempien maitten alueille.* "Norway and Russia have just agreed on a joint visa-free border zone for the territories of both countries."

> **Example 18** (HOPE, "foreign and security policy", 2020-11-25): *Pohjoismaiden ja Venäjän välisen yhteistyön osalta ulkoasiainvaliokunnassa korostettiin parlamentaarisen vuoropuhelun tärkeyttä*. "Regarding cooperation between the Nordic countries and Russia, the Foreign Affairs Committee emphasized the importance of parliamentary dialogue."

Topic "employment" showed a decline in SADN (**Figure 5d**); as this trend does not align with the fluctuation in Finland's unemployment rate during the observed period (Stat.fi 2025), the change in tone may instead point to evolving rhetorical conventions rather than underlying economic conditions. Based on close reading, most sentences labelled as SADN seemed to represent more the disappointment aspect of the category than sadness, exemplified in **Example 19** and **Example 20** that talk about slow or inadequate improvements in employment rates.

> **Example 19** (SADN, "employment", 2000-03-24): *Työllisyyden paraneminen maassamme on edelleen monin paikoin tuskastuttavan hidasta*. "The improvement in employment in our country is still painfully slow in many places."

> **Example 20** (SADN, "employment", 2020-11-24): *Me odotimme, että hallitus olisi täydentänyt talousarvioesityksen suurimmat puutteet eli täydentänyt nämä puuttuvat työllisyystoimet tähän hallituksen talousarvioesitykseen, mutta eihän niitä työllisyystoimia vieläkään näy eikä kuulu*. "We expected the government to have filled in the biggest gaps in the budget proposal, that is, to have filled in these missing employment measures in this government budget proposal, but such employment measures are still nowhere to be seen."

The only meaningful increasing trend was observed with "social and health care" within HATE (**Figure 5e**), suggesting that expressions of hate, disgust, taunts, or mockery have become more frequent when MPs discuss matters of social and health care. The most likely explanation for this development is the prolonged and much debated social and health care reform (*sote-uudistus)* that was discussed as early as 2011 but implemented in 2023. **Examples 21** and **22** express similar emotions towards opposing views of the same issue.

> **Example 21** (HATE, "social and health care", 2012-06-20): *[E]i voi olla niin, että ensin luodaan kriteerit vahvalle peruskunnalle ja viedään tätä kuntakentälle, yhtäkkiä tulee sitten joku sote-uudistus, joka kumoaa tämän ja pakottaa kuntia tai ajaa kuntia johonkin sote-malliin (…)*. "It cannot be that we first create criteria for a strong basic municipality and take this to the municipal level, then suddenly some social welfare reform comes along that overturns this and forces or drives municipalities to adopt some social welfare model."

**Example 22** (HATE, "social and health care", 2012-06-20): *Sote-uudistus ei ole jostain nurkan takaa tuleva mörkö, joka ryhtyy horjuttamaan kuntauudistuksen sinänsä erinomaisia hallitusohjelmaan kirjattuja tavoitteita.* "The social welfare reform is not a monster coming from around the corner that will undermine the excellent goals of the municipal reform, which are included in the government program."

In sum, this subsection described the main findings in the topic-emotion expressions over time. Five meaningful decreasing trends were observed, most importantly the "parliamentary factions" topic decreasing in all negative emotion categories and "democracy" declining in both HOPE and FEAR categories. Only one increasing trend was observed, that of "social and health care" in the HATE category.

## Discussion

In this final section, the results are discussed and interpreted further. Some limiting factors for the interpretations are also discussed in a separate subsection.

This paper studied the relationship between emotions and topics in Finnish parliamentary speech. By combining large-scale quantitative analysis with close reading, the analysis aimed to show how emotion expression is distributed across policy areas and how these patterns have evolved over time, effectively describing a *topic landscape*.

The underlying idea of a topic landscape "painting" invokes impressions of broad strokes and few details, as something is viewed from afar. This aligns with quantitative distant reading, where averages convey the overall structure and close reading is used to add recognizable details for depth. Considering the full corpus as "one point in time" (synchronic perspective) and examining the results of this study in increasing order of detail revealed a topic-emotion "painting" that would otherwise have been too expansive to grasp as a whole.

While negative emotions as a group dominated the landscape and gave the "painting" a generally negative "hue", the finer-grained topic-emotion findings suggest that specific emotion combinations tend to concentrate on certain thematic fields. The three largest emotion categories – NEUT, HOPE and FEAR – seem to dominate the topic landscape and even form a pattern of duality: the calm, neutral background of the institutional speech contrasted by the strongly polarized areas concerning "employment" and "energy". These main findings are also partly supported by earlier studies, specifically by Elo *et al.* (2025) who studied "economy" and "environmental politics", both closely related to issues regarding energy solutions as well as employment discussions. Widman (2025) also found that "energy" as a topic tends to evoke polarizing views.

Earlier studies have reported increased polarization in Finland (e.g. Suojanen, Lehtonen & Saarinen 2024) and listed multiple adverse effects of polarization for the society, such as weakening of relations between population groups, growth of hate speech and hate crimes, and growth of unwanted interpretations of strategic communication (*ibid*.: 3). Against that backdrop, it is noteworthy that the present analysis found only two topics that exhibited a clearly polarized emotional stance. This suggests that, within parliamentary speech, polarization may be concentrated within specific policy areas. Decisions made within polarized topics probably tend to have far-reaching effects, so it is natural for such discussions to become more emotional, as all available tools are being used to persuade the other side. Such topics may indicate more precisely where contentious societal cleavages are most salient, so "employment" and "energy" may serve as useful indicator topics for polarization dynamics in future research.

Negative skewness was most pronounced in topics about distributive politics ("budget" and "taxation") and when the MPs talked about the opposing parties ("parliamentary factions"). Competitive elements are inherent to politics, but it is notable that distributive policy areas are predominantly associated with negative emotions rather than polarization. While the allocation of resources can also generate positive outcomes, budgetary debates appear to be framed mainly in terms of insufficiency. Such framing resonates with neoliberal logics that prioritize efficiency and restraint, potentially narrowing the emotional and ideological space for positive narratives around public spending.

In the positive skewness group, most topics were overrepresented in HOPE, underlining the importance of this category. "Foreign and security policy" and "development cooperation" were also underrepresented in all negative emotion categories, which suggests that matters outside the domestic sphere may relate most strongly to positivity and to HOPE. This pattern may reflect the relative emotional distance of issues occurring outside the domestic sphere, which may make positive framing easier. However, as topics such as "public sector" and "agriculture" also appeared in this group, it is more likely that Finland's position as a relatively small and peripheral country merely causes references to other countries function as benchmarks or sources of best practices rather than as objects of contestation. More close reading would be required to verify this view.

Despite the synchronic landscape showing mainly negativity with a strong forward-looking aspect (HOPE), the diachronic observations showed that the Finnish parliamentary speech has become steadily less negative during the observed period. Examining the constituent emotions revealed that HOPE was the most important factor behind this observed change as well, as its proportion was also increasing over time. Also, after 2011 FEAR has been decreasing, showing only a jump during 2020 in response to the Covid-19 crisis. Again, HOPE and FEAR seem to be the main forces that shape the tone of the parliamentary discussion.

Superficial viewing of the observed developments in rising HOPE and declining FEAR could be seen as indicating softening of value orientations, but rethinking HOPE as a category suggests that this may not be the case. HOPE is a category that in parliamentary contexts may be more about legitimizing institutional solutions rather than expressing true positivity. The observed post-election increases in HOPE may associate with the new governments' willingness to show off their commitment and forward-looking attitude specifically to their voters, which would somewhat align with Osnabrügge *et al*. (2021).

It may also be that the hardening of societal values suggested by earlier research has created a preference for optimism while negative emotion expression is seen as unhelpful and unwanted, maybe even a sign of weakness – a neoliberal world prefers quick solutions and strong leadership rather than complaints and hesitation. On the other hand, increasing HOPE can also suggest a kind of "jam tomorrow" mentality: that even though the situation is difficult right now, with the proposed decisions and solutions there will be a change for the better in the near future, while it remains unclear if this near future will ever actually be reached.

Observing the changes in topic expressions within the emotion categories suggested some further possible explanations for the increasing HOPE and declining FEAR. "Parliamentary factions" exhibited a decline across all negative emotion categories, which may indicate a reduction in overt hostility toward other parties. Close reading provided some support for this interpretation, as the language used in later contexts appeared less overtly confrontational than in earlier periods. However, this trend does not necessarily imply greater consensus or softened underlying values. Rather, it may reflect a shift toward more formalized or strategically restrained modes of expression, potentially associated with a hardening of political positions that are communicated through more indirect or rhetorically controlled means. This also highlights a limitation of the modeling approach: while explicit negativity may decrease, subtler forms such as sarcasm, irony, or performative politeness are unlikely to be captured by the model.

"Democracy" showed decline in both HOPE and FEAR, suggesting decreasing polarization. Rather than indicating a diminished importance of democracy itself, this pattern may suggest that democracy is invoked less frequently as an explicitly evaluative or emotionally charged frame and is instead framed more neutrally. One possible explanation would be that political discussion has perhaps shifted away from value-oriented argumentation altogether and moved toward more pragmatic, technical, or efficiency-oriented framings of policy issues. Such a shift could be consistent with a broader hardening of political values, where foundational principles are treated as settled or instrumentalized, rather than actively debated and emotionally negotiated. In this sense, reduced emotional engagement with democracy may reflect not depoliticization, but a change in how democratic values are rhetorically mobilized.

The topic "question time" showed significant prevalence in love-compliments category (LOVE), which – paired with close reading – suggests that the public nature of question time may encourage performative politeness. This topic did not exhibit any meaningful changes over time, meaning it is a rather stable topic emotion-wise. The finding of Osnabrügge *et al.* (2021) that "emotive rhetoric is used primarily when politicians think they are addressing not only their fellow parliamentarians but also voters" (*ibid*.: 896) seems to apply here, as question times are publicly broadcast. The fact that the emotion levels of "question time" have not significantly changed suggests that the importance of question time as a way of connecting to voters has maintained its status despite the Internet age offering multiple other channels for performative communication. The performance of complimenting colleagues in this setting may function as a low-cost rhetorical strategy for signaling credibility and cooperative intent to external audiences.

In summary, these results illustrated how the parliamentary topic landscape was shaped by issue-specific sensitivities to emotions. Neutrality (NEUT), hopeful-optimistic-trust (HOPE), and fear-worry-distrust (FEAR) seemed to dominate the emotions across the board, but HOPE exhibited a slow but constant increase in proportion. 'Hope' remains an underexplored emotion in both psychological and economic research and could be a viable avenue for further research even in the field of linguistics. A recent study by Graham and Mujcic (2025) found that higher levels of hope in individuals are associated with improved economic and employment outcomes, better health, and greater resilience to adverse shocks. These individual-level benefits may also extend to the institutional sphere, suggesting that fostering hope could yield broader social and economic advantages. In this sense, increasing levels of HOPE can be seen as a development that deserves further attention. Topic landscape studies and examination of topics that react to HOPE and its counterpart FEAR, such as "employment" and "energy", would offer valuable windows into better understanding polarization and the evolving political focus in Finland.

**Limitations**

Besides topics, a register-based analysis would also have been possible, since all speeches were classified into formal categories of speech type. However, the register categories in *Eduskunta* are institutionally predefined, offering only limited and rather predictable variation. By contrast, topics provided a more dynamic and nuanced view of parliamentary discourse, since they cut across speech types and reflected the substantive issues that generated emotions and stance-taking.

Many of the diachronic topic–emotion trends that did not meet the criteria for statistical significance nevertheless showed downward trajectories. While these individual trends cannot be interpreted in isolation, considering them collectively suggests that some topics may have declined across multiple emotion categories, potentially indicating a decrease in their overall prevalence in parliamentary discourse. Topic prevalence as such was not directly examined in this study; however, based on earlier findings by Ristilä and Elo (2023), whose analysis covered the first decade (2000–2010) of the observation period used in this study, no major long-term increases or decreases in topic prevalences would be expected. This lends tentative support to the interpretation that the observed declines may reflect gradual shifts in rhetorical emphasis rather than substantive changes in the topic proportions.

Lastly, because the topics were relatively broad, it is also possible that the observed diachronic changes in topic-emotion pairs were partly due to changing "sub-topics", or changes in themes within topics. Different themes can evoke very different emotion expression, but studying the effects of sub-topics would have required further and more fine-grained topic modeling that was out of scope for this paper.

# APPENDIX A: Topics and their most important words

| Interpretation | Topic # | Top 10 words |
|---|---|---|
| commerce | 0 | *kauppa, yhtiö, yrittäjä, kilpailu, myydä, markkinat, tuote, ostaa, kuluttaja, ulko#maa*<br>'commerce', 'company', 'entrepreneur', 'competition', 'to sell', 'the market', 'product', 'to buy', 'consumer', 'foreign country' |
| social benefits | 1 | *nuori, perhe, tuki, vanhempi, opiskelija, vanhus, lapsi#perhe, elämä, koti, kuntoutus*<br>'young', 'family', 'support', 'parent', 'student', 'elderly person', 'family with children', 'life', 'home', 'rehabilitation' |
| energy | 2 | *energia, käyttö, sähkö, uusiutuva, auto, puu, osaaminen, ydin#voima, päästö, metsä*<br>'energy', 'use', 'electricity', 'renewable', 'car', 'wood', 'know-how', 'nuclear energy', 'emission', 'forest' |
| regionality | 3 | *maa#kunta, harmaa, kaupunki, asukas, alueellinen, seutu, elin#tarvike, pohjoinen, asua, alue#politiikka*<br>'county', 'grey', 'city', 'inhabitant', 'regional', 'region', 'foodstuff', 'northern', 'to live', 'regional policy' |
| education | 4 | *yli#opisto, koulutus, koulu, opetus, kieli, ammatillinen, opiskelija, opettaja, oppilas, nuori*<br>'university', 'education', 'school', 'teaching', 'language', 'vocational', 'student', 'teacher', 'student', 'young person' |
| pensions | 5 | *eläke, laki#aloite, aloite, lääke, kansan#eläke, henkilö, kuukausi, korvaus, epä#kohta, korjata*<br>'pension', 'legislative proposal', 'initiative', 'medicine', 'social insurance', 'person', 'month', 'reimbursement', 'defect', 'to fix' |
| crime | 6 | *poliisi, tieto, rikos, valvonta, viran#omainen, henkilö, laki#valio#kunta, kertomus, oikeus#ministeriö, tuomio#istuin*<br>'police', 'knowledge', 'crime', 'supervision', 'public authority', 'person', 'Standing Committee on Civil-Law Legislation', 'story', 'Ministry of Justice', 'court' |
| employment | 7 | *työn#tekijä, työ#paikka, työtön, työn#antaja, työ#elämä, työttömyys, palkka, työ#voima, nuori, työllisyys*<br>'employee', 'workplace', 'unemployed', 'employee', 'working life', 'unemployment', 'pay', 'workforce', 'young person', 'employment' |
| legislation | 8 | *lain#säädäntö, käytäntö, peruste, oikeus, laki#esitys, säätää, tarkoitus, käyttö, muuttaa, lausunto*<br>'legislation', 'convention', 'basis', 'justice', 'legislative proposal', 'to degree', 'meaning', 'use', 'to change', |

| | | | |
|---|---|---|---|
| | | | 'statement' |
| | traffic and transport | 9 | liikenne, määrä#raha, hanke, tie, joukko#liikenne, lisäys, päivä#järjestys, kunto, rata, pää#luokka 'traffic', 'budget', 'project', 'road', 'public transport', 'increase', 'agenda', 'condition', 'track', 'section' |
| | question time | 10 | pohja, talous#valio#kunta, yksityis#kohtainen, valtio#varain#valio#kunta, n:o, yleis#keskustelu, laki#ehdotus, sallia, ilmoittaa, debatti 'base', 'Commerce Committee', 'detailed', 'Finance Committee', 'number', 'general discussion', 'legislative proposal', 'to allow', 'to report', 'debate' |
| | administration | 11 | tehtävä, ministeriö, henkilö, siirtää, hallinto, henkilöstö, viesti, hoitaa, perustaa, järjestää 'task', 'ministry', 'person', 'to move', 'administration', 'personnel', 'message', 'to take care of', 'to establish', 'to arrange' |
| | public sector | 12 | palvelu, tulevaisuus, merkittävä, taso, toimen#pide, tarve, jatko, toimi, julkinen, toteuttaa 'service', 'future', 'significant', 'level', 'operation', 'need', 'continuation', 'office', 'public', 'execute' |
| | foreign and security policy | 13 | sopimus, kansain#välinen, kansallinen, yhteinen, turvallisuus, linjaus, toimija, selonteo*, selonte*, maailma 'agreement', 'international', 'national', 'mutual', 'security', 'definition of policy', 'operator', 'report*', 'report*', 'the world' |
| | parliamentary factions | 14 | kokoomus, linja, pää#ministeri, keskusta, vastuu     sosiali#demokraatti, politiikka, edus#kunta#ryhmä, poliittinen, ajaa 'National Coalition Party', 'policy', 'Prime Minister', 'Center Party', 'responsibility'        ,'Social Democratic Party', 'politics', 'parliamentary group', 'political', 'to drive' |
| | voting | 15 | lausuma#ehdotus, äänestys, tyhjä, kone, nuoriso#työ, ehdokas#lista, urheilu, vaali, veikkaus#voitto#vara, maksimi 'draft declaration',  voting', 'empty', 'machine', 'youth work', 'candidate list', 'sports', 'election', 'profits from Veikkaus', 'maximum' |
| | law proposals | 16 | ehdottaa, vasta#lause, päättää, laki#ehdotus, hylätä, ensimmäinen, äänestää, mukainen, sisältö, pykälä 'to suggest', 'objection', 'to decide', 'legislative proposal', 'to discard', 'the first', 'to vote', 'consistent with', 'content', 'section' |
| | democracy | 17 | perustus#laki, presidentti, perustus#laki#valio#kunta, jäsen, valtio#neuvosto, valita, vaalit, tasa#valta, demokratia, valta 'constitution', 'president', 'Constitutional Law Committee', 'member', 'Government', 'to choose', 'election', 'republic', 'democracy', 'power' |
| | development | 18 | neuvosto, kulttuuri, jäsen#maa, yhteis#työ, järjestö, ase, kansain#välinen, kokous, pohjois#mainen, |

| | | |
|---|---|---|
| cooperation | | kehitys#yhteis#työ 'council', 'culture', 'member state', 'cooperation', 'organization', 'weapon', 'international', 'meeting', 'Nordic', 'development cooperation' |
| agriculture | 19 | maa#seutu, maa#talous, ruoka, tuki, tila, viljelijä, tuotanto, maa#talous#politiikka, metsä#talous, pelto 'rural area', 'agriculture', 'food', 'support', 'estate', 'farmer', 'production', 'agricultural policy', 'forestry', 'field' |
| social and health care | 20 | sosiaali, terveyden#huolto, palvelu, hoito, lähettää, lääkäri, ehdottaa, potilas, puhe#mies#neuvosto, terveys 'social', 'health care', 'service', 'care', 'to send', 'doctor', 'to suggest', 'patient', 'Speaker of Parliament Bureau', 'health' |
| taxation | 21 | verotus, vero, korotus, pieni#tuloinen, eläkeläinen, maksu, tulo, korottaa, nostaa, indeksi 'taxation', 'tax', 'increase', 'person with low income', 'pensioner', 'payment', 'income', 'to increase', 'to raise', 'index' |
| GENERAL | 22 | kysyä, puoli, ymmärtää, päästä, kertoa, paikka, uskoa, iso, ajatella, viedä 'to ask', 'half', 'to understand', 'to get to', 'to tell', 'place', 'believe', 'large', 'to think', 'to bring' |
| housing | 23 | pankki, asunto, laina, kiinteistö, innovaatio, asuminen, hinta, korko, rakentaminen, asua 'bank', 'apartment', 'loan', 'estate', 'innovation', 'living', 'price', 'interest', 'construction', 'live' |
| social problems | 24 | nainen, mies, alkoholi, nuori, tasa#arvo, väki#valta, isä, suku#puoli, liitto, elämä 'woman', 'man', 'alcohol', 'young person', 'equality', 'violence', 'father', 'gender', 'union', 'life' |
| budget | 25 | talous, budjetti, ensi, kasvu, meno, kasvaa, velka, työllisyys, lama, valtion#talous 'economy', 'budget', 'next', 'growth', 'expense', 'to grow', 'debt', 'employment', 'economic depression' |

# APPENDIX B: Topic prevalences in emotion subcorpora

Each column lists the topic prevalences in an emotion subcorpus, except for the last column which lists the average prevalences for each topic. These numbers were used to calculate the percentages in Table 1.

|  | HATE | SADN | FEAR | NEGA | NEUT | LOVE | HOPE | JOY- | POSI | average |
|---|---|---|---|---|---|---|---|---|---|---|
| **Commerce** | 0.01439 | 0.01136 | 0.01622 | 0.00669 | 0.01368 | 0.00498 | 0.0163 | 0.01032 | 0.01242 | **0.01182** |
| **Social benefits** | 0.01869 | 0.04944 | 0.04896 | 0.01518 | 0.03038 | 0.02167 | 0.05126 | 0.02925 | 0.04597 | **0.03453** |
| **Energy** | 0.01588 | 0.01227 | 0.03356 | 0.01181 | 0.01376 | 0.0099 | 0.03941 | 0.01307 | 0.01175 | **0.01793** |
| **Regionality** | 0.00689 | 0.00988 | 0.01005 | 0.00677 | 0.00752 | 0.00735 | 0.01046 | 0.01018 | 0.01553 | **0.00940** |
| **Education** | 0.01067 | 0.01555 | 0.01255 | 0.00947 | 0.0184 | 0.01625 | 0.0221 | 0.0217 | 0.03537 | **0.01801** |
| **Pensions** | 0.01188 | 0.01435 | 0.01448 | 0.00822 | 0.02674 | 0.02104 | 0.00849 | 0.0115 | 0.00666 | **0.01371** |
| **Crime** | 0.01586 | 0.02257 | 0.01728 | 0.01108 | 0.02596 | 0.01367 | 0.00941 | 0.01189 | 0.00789 | **0.01507** |
| **Employment** | 0.02238 | 0.03262 | 0.0361 | 0.01708 | 0.01934 | 0.01349 | 0.03583 | 0.02384 | 0.00048 | **0.02235** |
| **Legislation** | 0.07341 | 0.06934 | 0.06834 | 0.06341 | 0.17301 | 0.06456 | 0.04936 | 0.04961 | 0.04845 | **0.07328** |
| **Traffic and transport** | 0.00629 | 0.0099 | 0.00866 | 0.00492 | 0.01878 | 0.00721 | 0.0102 | 0.01018 | 0.00388 | **0.00889** |
| **Question time** | 0.00348 | 0.00284 | 0.00212 | 0.00426 | 0.0083 | 0.00729 | 0.00159 | 0.00424 | 0.0021 | **0.00402** |
| **Administration** | 0.0504 | 0.05112 | 0.04714 | 0.04954 | 0.06556 | 0.05352 | 0.0459 | 0.05317 | 0.07704 | **0.05482** |
| **Public sector** | 0.10974 | 0.16549 | 0.22017 | 0.092 | 0.16327 | 0.15286 | 0.32037 | 0.24064 | 0.04972 | **0.16825** |
| **Foreign and security policy** | 0.02038 | 0.01955 | 0.02648 | 0.01733 | 0.03176 | 0.02961 | 0.04234 | 0.03796 | 0.02242 | **0.02754** |
| **Parliamentary factions** | 0.11898 | 0.0989 | 0.0517 | 0.15614 | 0.02831 | 0.07898 | 0.03814 | 0.06553 | 0.05912 | **0.07731** |
| **Voting** | 0.003 | 0.00319 | 0.00202 | 0.00357 | 0.00477 | 0.00466 | 0.00189 | 0.00313 | 0.00507 | **0.00348** |
| **Law proposals** | 0.00765 | 0.00372 | 0.00321 | 0.00717 | 0.03198 | 0.00434 | 0.00259 | 0.00395 | 0.00027 | **0.00721** |
| **Democracy** | 0.03946 | 0.02587 | 0.02016 | 0.03601 | 0.04099 | 0.04251 | 0.02481 | 0.03193 | 0.01776 | **0.03106** |
| **Development cooperation** | 0.007 | 0.00736 | 0.00666 | 0.00578 | 0.0106 | 0.01478 | 0.0101 | 0.01733 | 0.0161 | **0.01063** |
| **Agriculture** | 0.00458 | 0.00714 | 0.00779 | 0.00387 | 0.00568 | 0.00519 | 0.00871 | 0.00627 | 0.00244 | **0.00574** |
| **Social and health care** | 0.01252 | 0.0131 | 0.01264 | 0.00834 | 0.02111 | 0.01006 | 0.01258 | 0.00998 | 0.00549 | **0.01176** |
| **Taxation** | 0.02635 | 0.02216 | 0.02327 | 0.01499 | 0.02217 | 0.00684 | 0.01369 | 0.01202 | 0.00803 | **0.01661** |
| **GENERAL** | 0.3346 | 0.25891 | 0.21304 | 0.40053 | 0.16432 | 0.3745 | 0.16431 | 0.25676 | 0.43914 | **0.28957** |
| **Housing** | 0.02962 | 0.02564 | 0.03004 | 0.01849 | 0.0188 | 0.01283 | 0.02035 | 0.0211 | 0.05523 | **0.02579** |
| **Social problems** | 0.01207 | 0.01267 | 0.01309 | 0.01045 | 0.00899 | 0.0104 | 0.00752 | 0.00889 | 0.04012 | **0.01380** |
| **Budget** | 0.0238 | 0.03505 | 0.0543 | 0.01689 | 0.02583 | 0.01151 | 0.03231 | 0.03556 | 0.01152 | **0.02742** |

# APPENDIX C: Monthly three-month emotions in topics

Graphs that are statistically significant (p<0.05) are on a white background, while not statistically significant graphs are on a grey background.

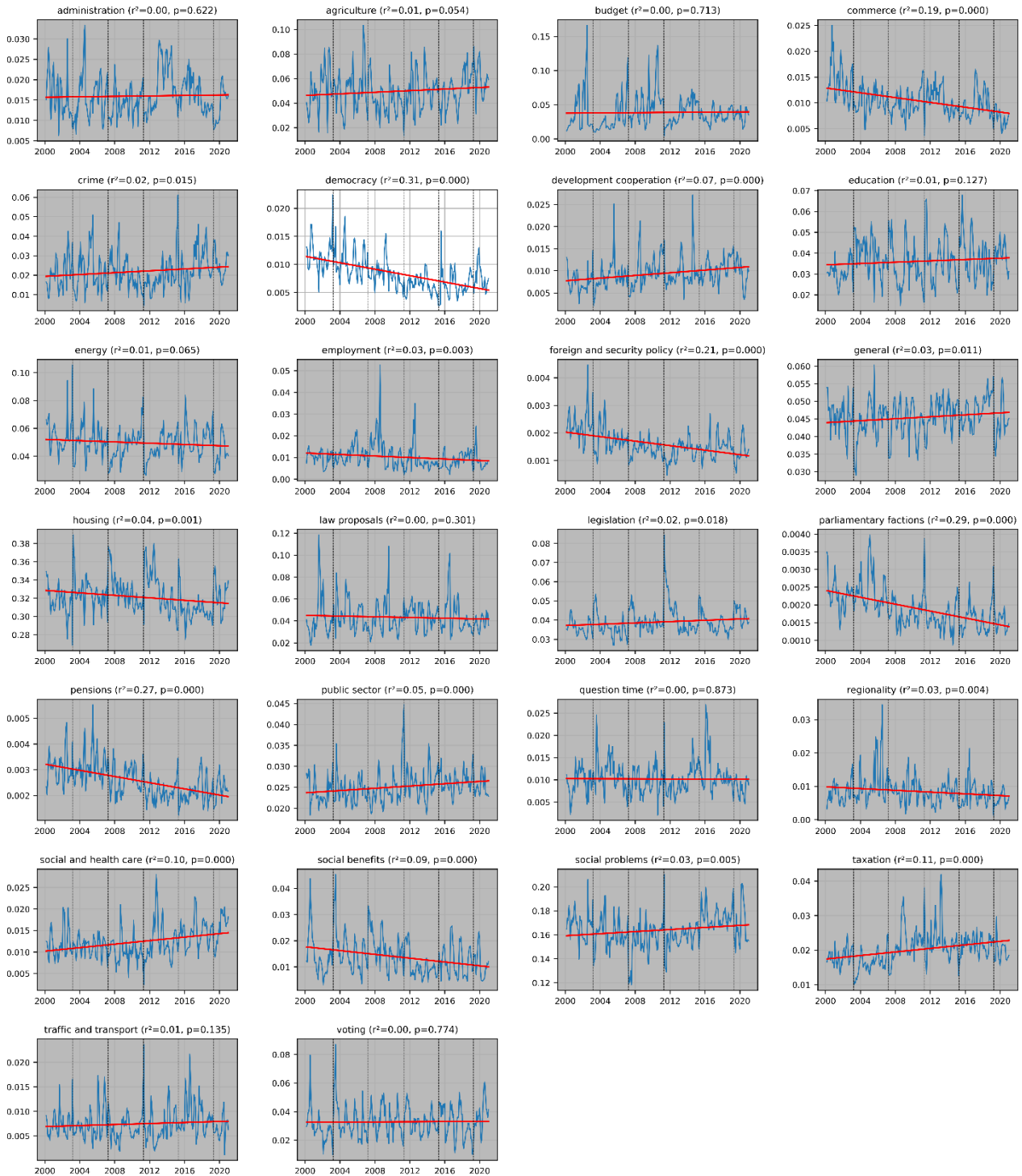

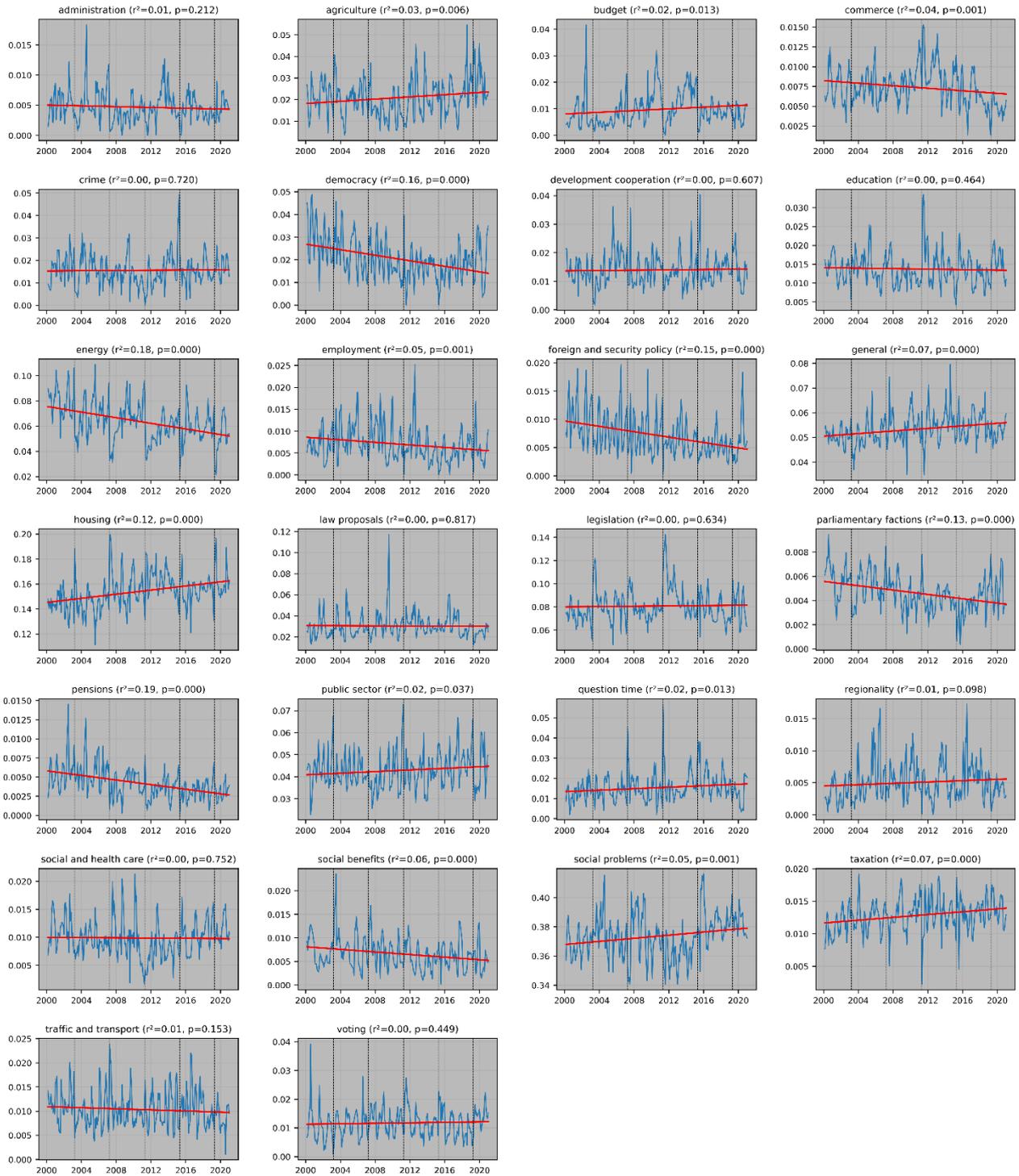

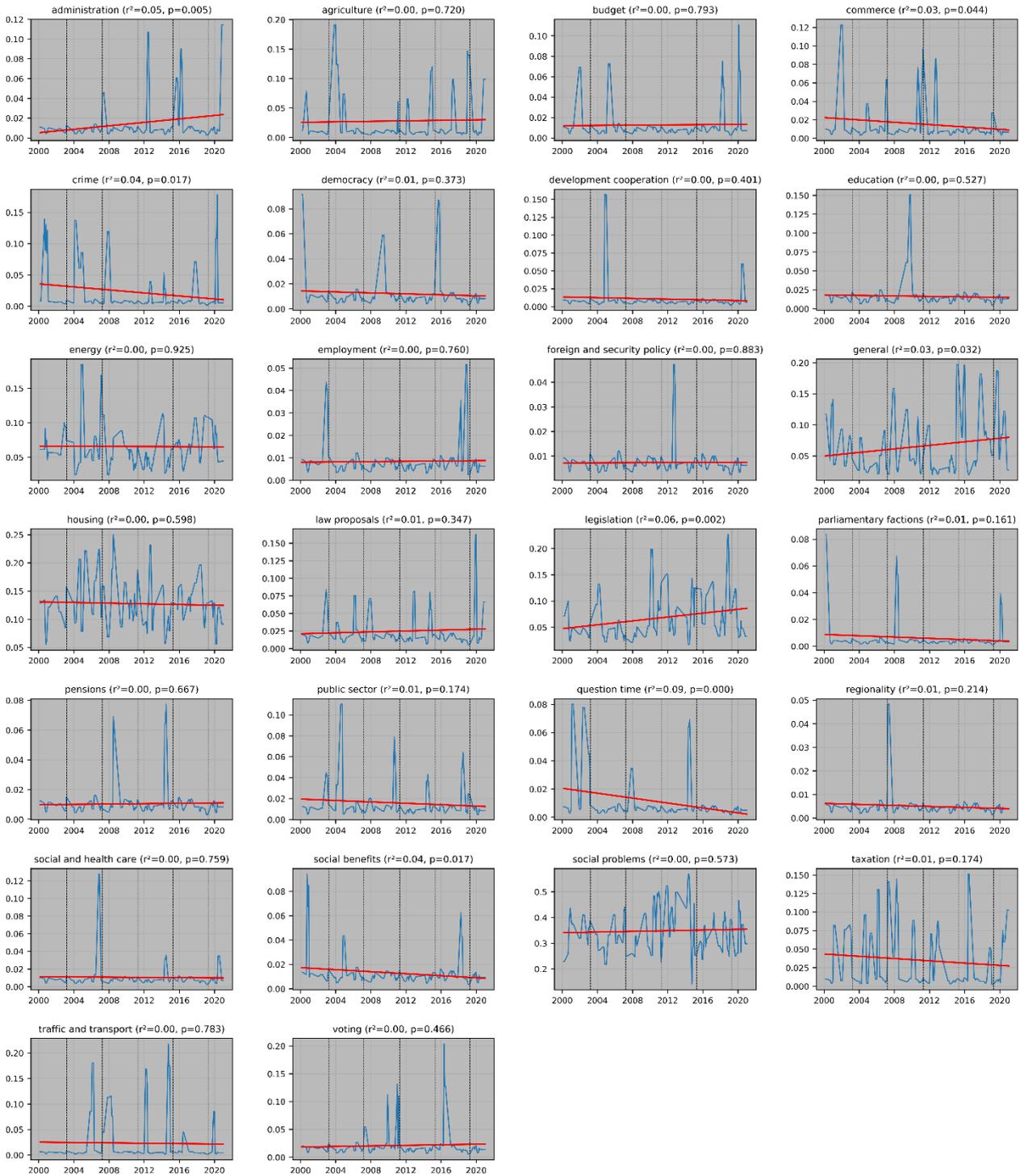

# neutral — 3-Month Rolling Average by Topic

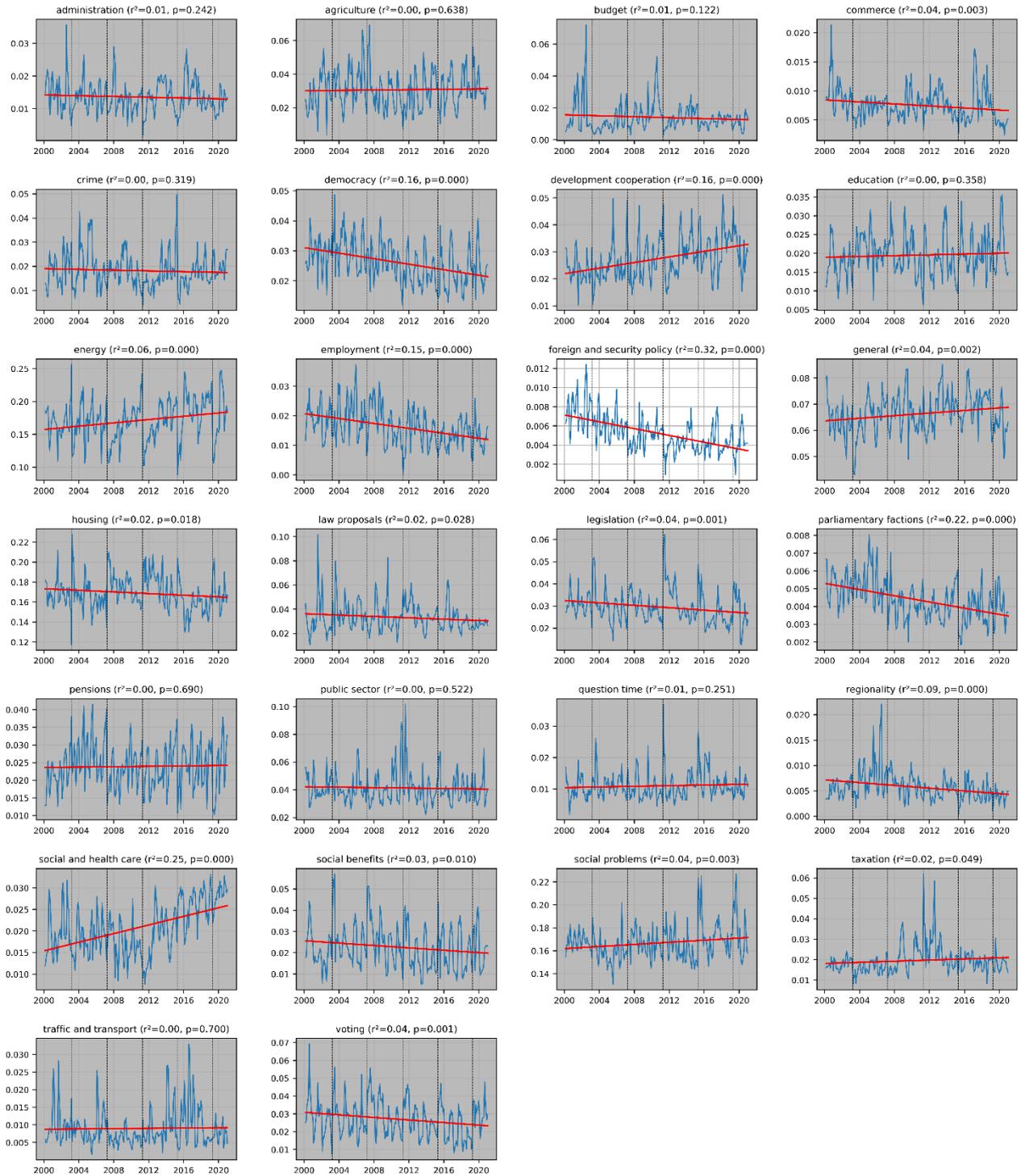

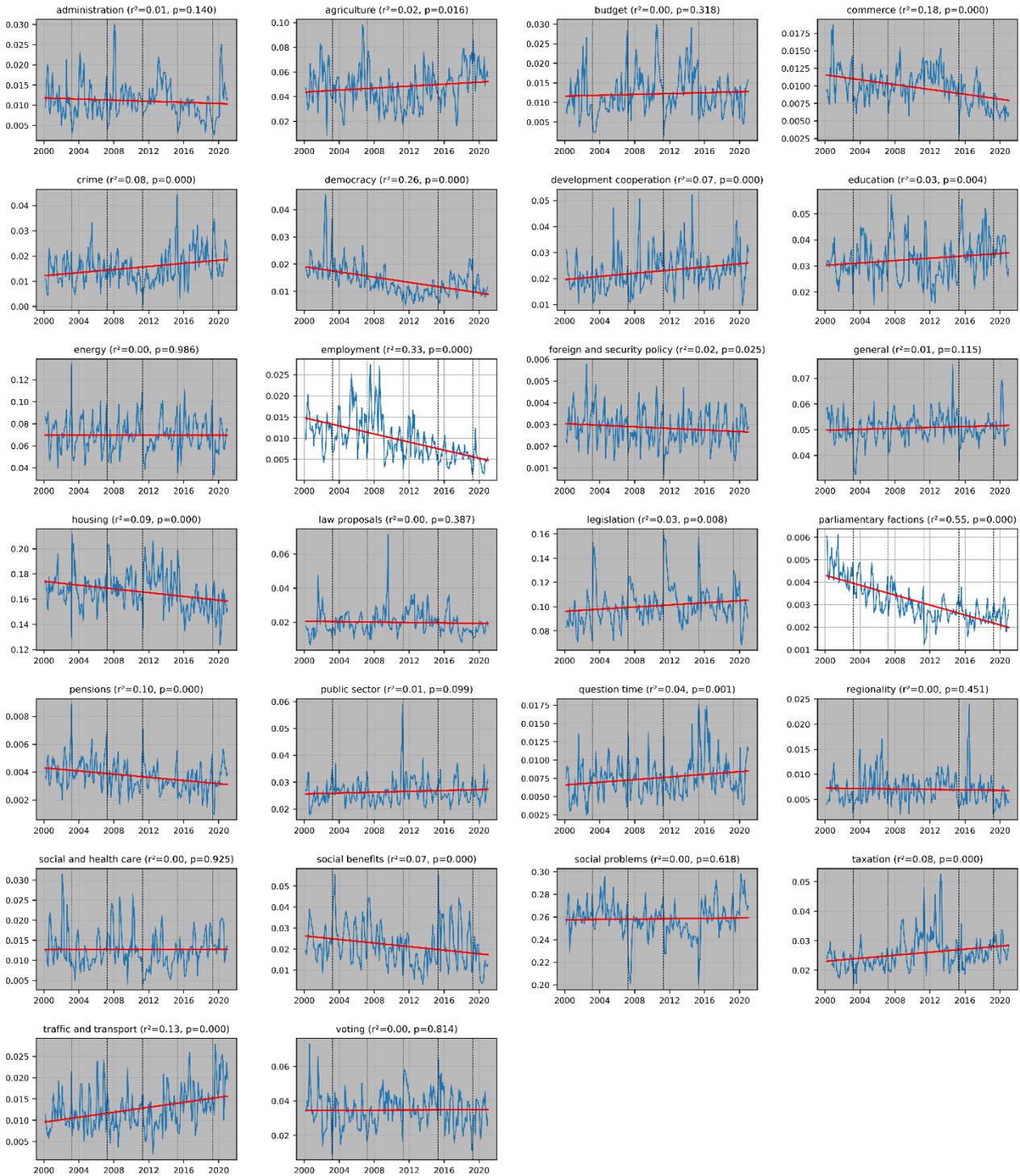

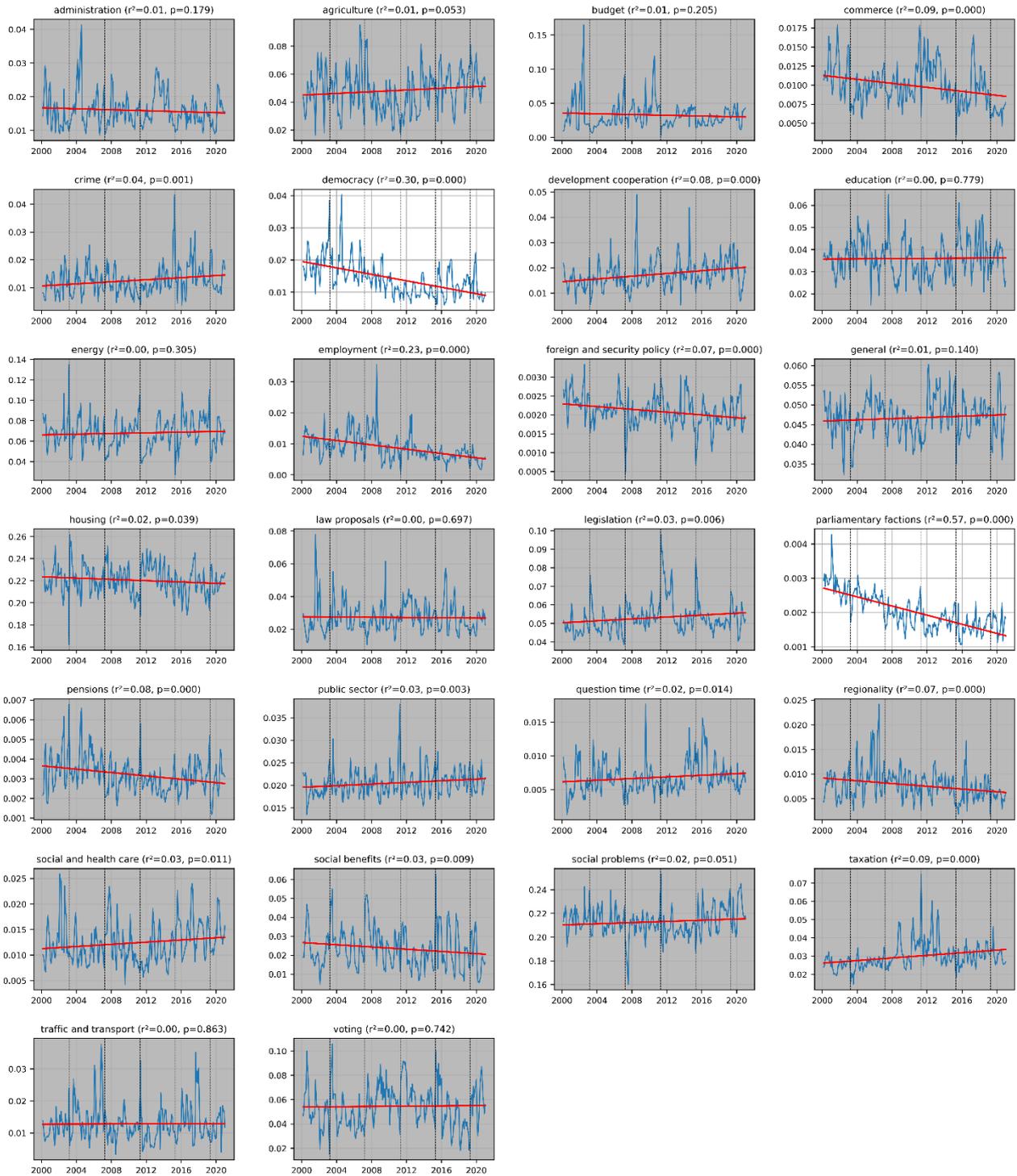

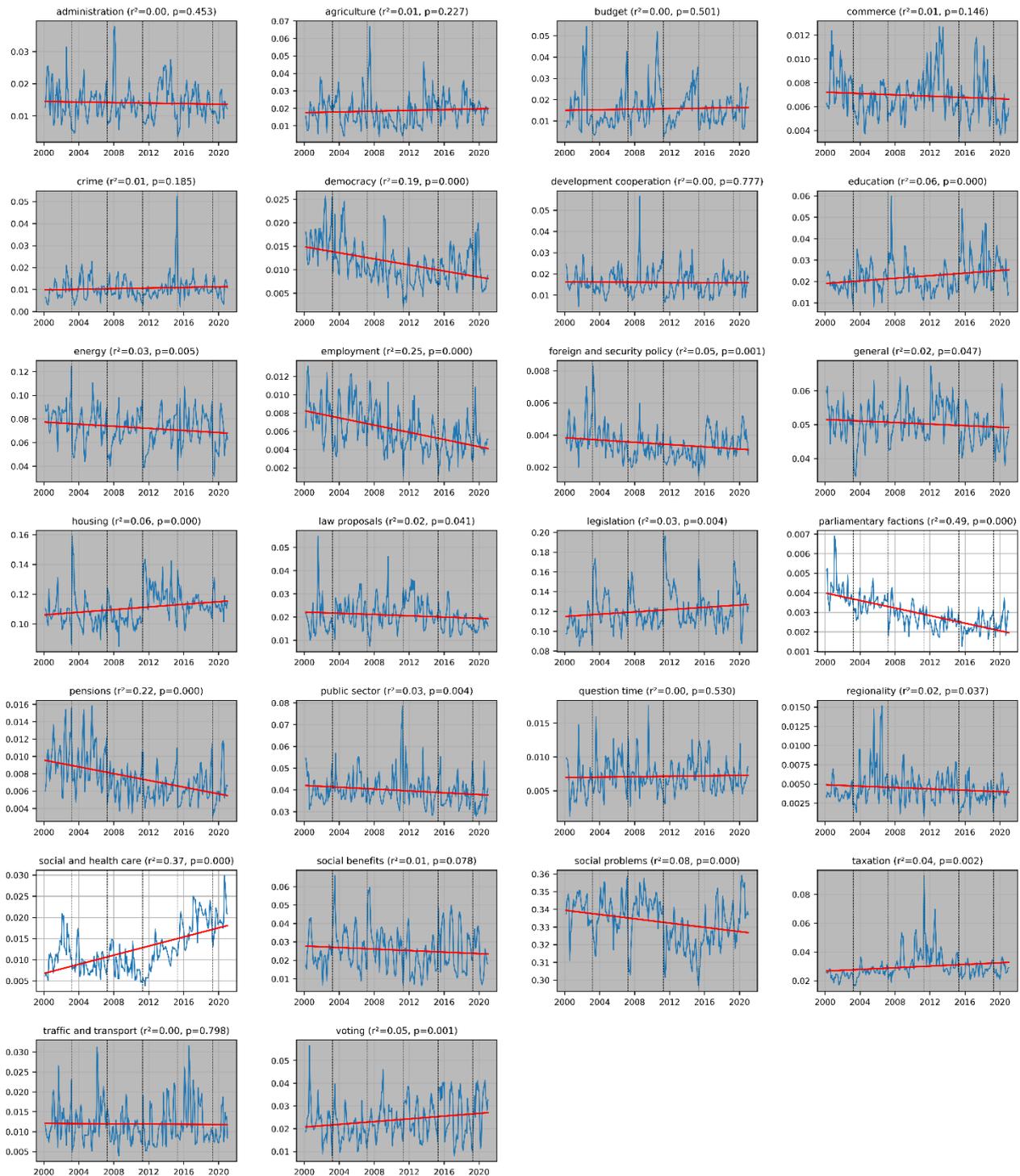

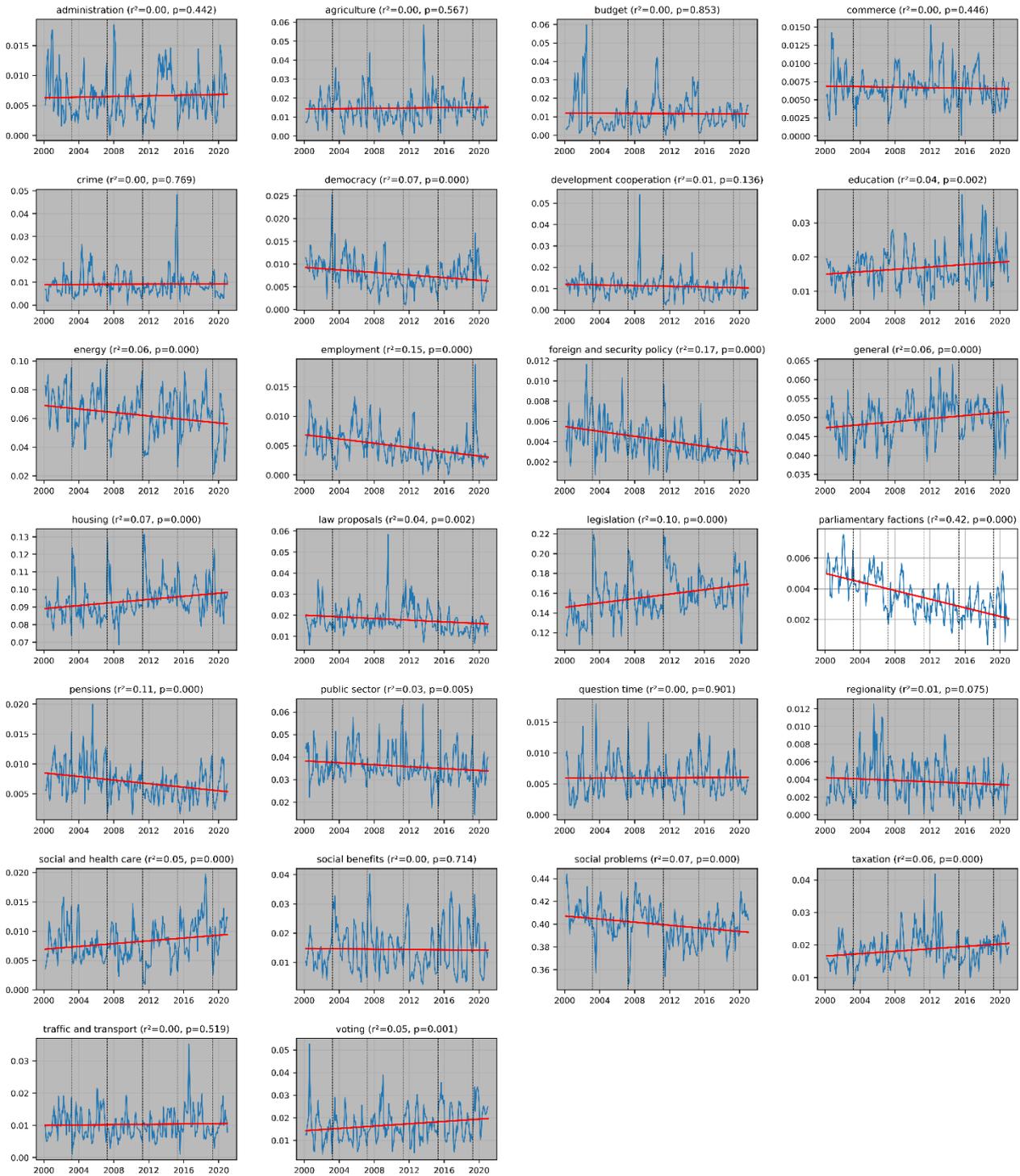

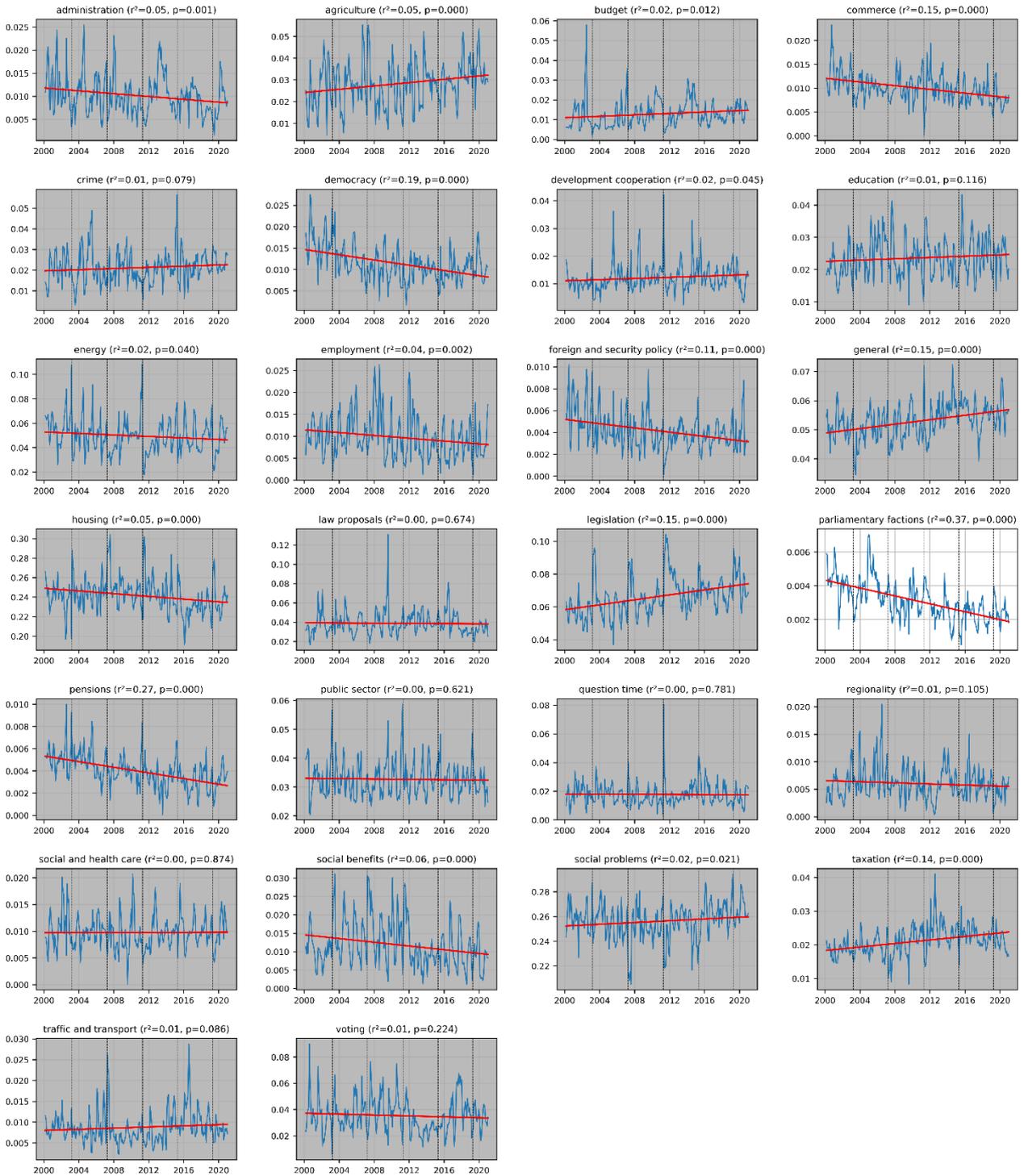